\documentclass{article}
\usepackage[utf8]{inputenc}

\PassOptionsToPackage{numbers, compress}{natbib}
\usepackage[final]{neurips_2025}

\usepackage[T1]{fontenc}
\usepackage{times}
\usepackage{amsmath, amssymb, amsthm}
\usepackage{mathtools}
\usepackage{bbm}
\usepackage{dsfont}

\usepackage[dvipsnames]{xcolor}
\definecolor{darkpink}{RGB}{199,21,140}

\usepackage{graphicx}

\usepackage{booktabs, array}

\usepackage{listings}
\usepackage{fancyvrb}
\fvset{fontsize=\small}

\definecolor{citecolor}{RGB}{0,102,204}
\definecolor{linkcolor}{RGB}{190,105,30}
\definecolor{urlcolor}{RGB}{199,21,133}

\usepackage{silence}
\usepackage[colorlinks,linktoc=all]{hyperref}
\usepackage[all]{hypcap}
\hypersetup{citecolor=citecolor}
\hypersetup{linkcolor=linkcolor}
\hypersetup{urlcolor=urlcolor}
\usepackage[nameinlink,capitalise]{cleveref}
\creflabelformat{equation}{#2\textup{#1}#3}  
\crefname{section}{\S}{\S\S}

\lstdefinestyle{mystyle}{
    commentstyle=\color{OliveGreen},
    numberstyle=\tiny\color{black!60},
    stringstyle=\color{BrickRed},
    basicstyle=\ttfamily\scriptsize,
    breakatwhitespace=false,
    breaklines=true,
    captionpos=b,
    keepspaces=true,
    numbers=none,
    numbersep=5pt,
    showspaces=false,
    showstringspaces=false,
    showtabs=false,
    tabsize=2
}
\newcommand{\bx}{\mathbf{x}}

\newcommand{\bse}{\boldsymbol{e}}

\newcommand{\bsh}{\boldsymbol{h}}

\newcommand{\bsm}{\boldsymbol{m}}

\newcommand{\bsq}{\boldsymbol{q}}

\newcommand{\bsx}{\boldsymbol{x}}
\newcommand{\bsy}{\boldsymbol{y}}

\newcommand{\bsD}{\boldsymbol{D}}

\newcommand{\bsH}{\boldsymbol{H}}

\newcommand{\bsW}{\boldsymbol{W}}

\newcommand{\calL}{{\mathcal{L}}}

\newcommand{\calX}{{\mathcal{X}}}
\newcommand{\calY}{{\mathcal{Y}}}

\newcommand{\bbR}{\mathbb{R}}

\newcommand{\bPi}{\boldsymbol{\Pi}}

\theoremstyle{plain}
\newtheorem{thm}{Theorem}[section]

\theoremstyle{definition}

\theoremstyle{remark}

\newtheorem{assm}{Assumption}

\newcommand{\norm}[1]{\left\Vert{#1}\right\Vert}

\def\[#1\]{\begin{equation}\begin{aligned}#1\end{aligned}\end{equation}}

\newcommand{\Tr}[2]{\text{T}_{#1 #2}}

\usepackage{tikz}
\usetikzlibrary{graphs}
\usepackage{wrapfig}
\usepackage{subcaption}
\usepackage{siunitx}
\usepackage{textcomp}
\usepackage{float}

\newsavebox\CBox

\def\UL#1{\underline{#1}}

\newcommand{\Times}{\!\times\!}

\title{Axial Neural Networks for Dimension-Free Foundation Models}
\author{
  Hyunsu Kim\textsuperscript{\textdagger}, Jonggeon Park\textsuperscript{\textdagger}, Joan Bruna\textsuperscript{\textdaggerdbl}, Hongseok Yang\textsuperscript{\textdagger}, Juho Lee\textsuperscript{\textdagger}\\
  \textsuperscript{\textdagger}KAIST, \textsuperscript{\textdaggerdbl}New York University,\\
  \texttt{\{kim.hyunsu,parkjonggeon,hongseok.yang,juholee\}@kaist.ac.kr}\\
  \texttt{bruna@cims.nyu.edu}
}

\begin{document}

\maketitle

\begin{abstract}
The advent of foundation models in AI has significantly advanced general-purpose learning, enabling remarkable capabilities in zero-shot inference and in-context learning.
However, training such models on physics data, including solutions to partial differential equations (PDEs), poses a unique challenge due to varying dimensionalities across different systems. 
Traditional approaches either fix a maximum dimension or employ separate encoders for different dimensionalities, resulting in inefficiencies.
To address this, we propose a dimension-agnostic neural network architecture, the Axial Neural Network (XNN), inspired by parameter-sharing structures such as Deep Sets and Graph Neural Networks. 
XNN generalizes across varying tensor dimensions while maintaining computational efficiency. 
We convert existing PDE foundation models into axial neural networks and evaluate their performance across three training scenarios: training from scratch, pretraining on multiple PDEs, and fine-tuning on a single PDE. 
Our experiments show that XNNs perform competitively with original models and exhibit superior generalization to unseen dimensions, highlighting the importance of multidimensional pretraining for foundation models.
\end{abstract}


\section{Introduction}
\label{sec:main:introduction}

The growing scale of deep learning models has led to the emergence of general-purpose AI systems, often called \emph{foundation models}. 
Trained over a large amount of unlabeled data with self supervision, these models have shown impressive generalization performance, enabling effective zero-shot inference and in-context learning on a wide range of tasks. In practice, these models are further fine-tuned or post-trained for particular target tasks, 
achieving performance superior to models trained from scratch.
A key requirement for developing highly-performing foundation models is the use of vast and diverse training data.
In fact, the relationship between a model's performance and the scale of data (and models) is known to follow a version of power law, called scaling law~\citep{kaplan2020scaling,havrilla2024understanding,bordelon2024dynamical}.

This paper is concerned with developing key techniques for building successful foundation models for physics data, such as climate time-series data and solutions of partial differential equations (PDEs). When training such a model, we often have to combine datasets from multiple systems or differential equations that operate on different dimensionalities. 
For instance, the Burgers equation describing dissipative fluid flow is usually studied in one spatial dimension, whereas the Navier–Stokes equations are studied in two or three dimensions. Since the solutions of these equations are typically represented as tensors whose elements are points in spatial and temporal grids, different dimensionalities mean that those tensors storing solutions have different numbers of axes.

A straightforward way to address such mixed-dimensional scenarios is to fix a maximum dimension and either pad lower-dimensional inputs with zeros or build separate encoders for different dimensions that share the same output space. 
However, both approaches are inefficient for low-dimensional data and incapable of handling inputs whose dimension exceeds the fixed maximum.
Consequently, most of the prior works on PDEs have developed models tailored to a specific dimension (typically 2D)~\citep{hao2024dpot,mccabe2024multiple,wang2024cvit,herde2024poseidon}. Extending these models to other dimensions is nontrivial. For instance, the commonly-used patchify operation, which applies 2D convolution to extract 16×16 patches for Transformer models~\citep{dosovitskiy2021image}, is inherently limited to 2D inputs. Although a patchify operation can be designed for any specific dimension, a 2D patchifier is only applicable to 2D inputs; processing 1D or 3D data requires an entirely different set of parameters, as these models lack an intrinsic parameter-sharing mechanism across dimensions.
While recent work has proposed dimension-agnostic methods based on neural processes~\citep{lee2025dimension} and multilayer perceptrons~\citep{levin2024any}, these methods still incur large computational costs: they either flatten high-dimensional data and process long sequences, leading to expensive attention layers, or pre-compute dimension-equivariant weights via singular-value decomposition, which becomes infeasible in high dimensions.

We propose an efficient, dimension-agnostic, and dimension-generalizable neural network by adopting the core principle of parameter-sharing from permutation-equivariant architectures such as Deep Sets~\citep{zaheer2017deep} and Graph Neural Networks~\citep{sperduti1997supervised,gori2005new,scarselli2009graph}, closely related to De Finetti’s theorem~\citep{finetti1931funzione,zaheer2017deep,bloem-reddy2020probabilistic}.
Our architecture achieves permutation equivariance over tensor axes. 
Concretely, we treat the axes of a tensor as elements of a \emph{set} and introduce a permutation-equivariant architecture which we refer to as the Axial Neural Network (XNN). 
Although such set-based XNN is simple and computationally efficient, we find it inherently less expressive, and thus further propose an advanced version termed graph-based XNN, which captures relationships among axes by treating the axes of a tensor as vertices of a graph.
Finally, we introduce a dimension-agnostic PDE foundational model trained and evaluated on PDEs of varying dimensionality within a single model. Crucially, and in contrast to traditional patchify operations, an XNN-based patchify operation leverages this parameter sharing to make it applicable to inputs of any dimension without modification.

To demonstrate the expressivity and benefits of multidimensional pretraining, we convert existing PDE foundation models~\citep{mccabe2024multiple, wang2024cvit} to the variants based on our XNNs. 
We evaluate the resulting models in three different settings: training a single PDE from scratch, pretraining with multiple PDEs, and fine-tuning on a single PDE. 
We show that our variants perform competitively with their original counterparts. 
We also conduct experiments to demonstrate the unseen-dimension generalization ability of XNNs, which plays an important role in such a dimension-agnostic strategy. 
Our XNN architecture shows better performance in unseen dimension fine-tuning, which underscores the necessity of multidimensional pretraining for foundation models. The implemented architectures are summarized in \url{https://github.com/kim-hyunsu/XNN}.
\section{Backgrounds}
\label{sec:main:background}

\subsection{Graph Neural Networks and Deep Sets}
A \textbf{Graph Neural Network (GNN)} is a deep neural network designed to process and make predictions on data represented as a graph~\citep{scarselli2009graph,zhou2020graph}. GNNs are characterized by a message-passing mechanism, in which information is exchanged between nodes through their connections, called edges. Formally, given the feature vector $\bsx_a$ of node $a$, its hidden representation $\bsh_a$ is computed as
\[
\label{eq:gnn}
\bsh_a = \phi\bigg(\bsx_a,\; \bigoplus_{b \in \operatorname{ngbr}(a)} \psi(\bsx_a, \bsx_b, \bse_{ab})\bigg),
\]
where $\phi$ and $\psi$ are neural networks with parameters shared across all nodes, $\bigoplus$ denotes a permutation-invariant aggregation operation, $\operatorname{ngbr}(a)$ is the set of neighbors of node $a$, and $\bse_{ab}$ is the edge feature between nodes $a$ and $b$. The critical architectural feature of a GNN is that the parameters of $\phi$ and $\psi$ are shared across all nodes in the graph. This means the exact same functions are used to update each node's representation based on its local neighborhood.

This parameter-sharing structure is the fundamental reason GNNs are permutation-equivariant: permuting the input nodes simply changes the order of identical operations, leading to a corresponding permutation in the output. This symmetry is therefore structurally embedded in the model design, allowing the GNN to generalize across graphs of different sizes and structures.

A \textbf{Deep Set} is a neural network for set-structured data \citep{zaheer2017deep} and can be viewed as a special case of GNNs in which every node is equally connected to every other node. They can therefore be expressed by the following simplification of \cref{eq:gnn}:
\[
\label{eq:deepset}
\bsh = \phi\bigg(\sum_{i=1}^{K} \psi(\bsx_i)\bigg),
\]
where $K$ is the number of elements in the set. Similar to GNNs, the design relies on parameter sharing: the same function $\psi$ is applied to every element $\bsx_i$ before a permutation-invariant aggregation (summation or maximum) is performed. This shared application of $\psi$ ensures the model is permutation-invariant by construction.

\subsection{Transpose and Axis-Permutation Equivariance}
The transpose of a matrix flips the matrix over its diagonal, and the transpose of a tensor swaps two axis indices.  Exchanging axes $i,j\in[1,K]$ of a rank-$K$ tensor $\bsx = (x_{d_1 \cdots d_K})_{d_1,\ldots,d_k}$ yields
\[
\bsx^{\top_{ij}}_{d_1 \cdots d_i \cdots d_j \cdots d_K}
    := x_{d_1 \cdots d_j \cdots d_i \cdots d_K},
\]
where $\top_{ij}$ denotes the transpose between axes $i$ and $j$. An axis permutation is obtained by cumulative transposes, corresponding to reordering the axes. For a rank-4 tensor $\bsx$, for example,
\[
\Pi(\bsx)_{d_1 d_2 d_3 d_4} =x_{\,d_{\pi(1)}\,d_{\pi(2)}\,d_{\pi(3)}\,d_{\pi(4)}}
       =x_{d_3 d_2 d_4 d_1},
\]
where $\pi$ is the permutation associated with $\Pi$.

Equivariance is the property that a function commutes with the action of a symmetry group $G$.  If $\calX$ and $\calY$ are acted on by \(\rho_\calX(g)\) and \(\rho_\calY(g)\) for each group element $g \in G$, respectively, then \(\phi:\calX\!\to\!\calY\) is \(G\)-equivariant when
\[
\phi\bigl(\rho_\calX(g)\,\bsx\bigr)=\rho_\calY(g)\,\phi(\bsx),
\quad\forall g\in G.
\]
Let \(\bPi\) be the group of all axis permutations of a rank-\(K\) tensor.  A mapping \(\phi\) is \emph{axis-permutation equivariant} if
\[
\label{eq:axis-perm-equiv}
\phi\bigl(\Pi(\bsx)\bigr)=\Pi\bigl(\phi(\bsx)\bigr),
\quad\forall\Pi\in\bPi,
\]
i.e., permuting the input axes and then applying \(\phi\) produces the same result as applying \(\phi\) first and then permuting the output.

\subsection{Cycle Notation for Axes Permutation}
\label{sec:main:cycle}
To express the reordering of axes throughout this paper, we require a precise notation for permutations. A permutation is formally defined as a bijection (a one-to-one mapping) from a set onto itself. In our context, the set consists of axis indices, such as $\{H, W, D\}$ for $\bx\in\bbR^{H\Times W\Times D}$.

First, we review the standard \emph{cycle notation} common in algebra. This notation uses parentheses $()$ to group elements into disjoint cycles that show the path each element follows under the permutation. An element within a cycle is mapped to the element immediately following it. The last element in a cycle is mapped back to the first, completing the loop.

For example, consider permutations on the set of three axis indices $\{1, 2, 3\}$:
\begin{itemize}
\item A permutation that maps $1 \to 3$, $3 \to 2$, and $2 \to 1$ is written as the single cycle $(1 \ 3 \ 2)$.
\item A permutation that maps $1 \to 3$, $3 \to 1$, and leaves 2 unchanged, $2 \to 2$, is written as $(1 \ 3)$. The element 2 is a fixed point.
\end{itemize}

Adopting this cycle notation, we define the transformation $\Tr{}{i_1 i_2 \dots i_n}$ as the permutation that maps the original ordered set of axes $d_1\Times d_2\Times \cdots \Times d_n$ to the new ordered arrangement $d_{i_1}\Times d_{i_2}\Times\cdots\Times d_{i_n}$. For example,
\[
&\bsy = \Tr{132}{}(\bsx),\quad\bsy\in\bbR^{W\Times D\Times H},\quad\bsx\in\bbR^{H\Times W\Times D},\\
&\bsy' = \Tr{13}{}(\bsx),\quad\bsy'\in\bbR^{D\Times W\Times H},\quad\bsx\in\bbR^{H\Times W\Times D}.
\]
\section{Axial Neural Networks}
\label{sec:main:method}

We draw inspiration from permutation-equivariant architectures such as Deep Sets~\citep{zaheer2017deep} and GNNs, which process set or graph data with a variable number of elements. A key advantage of these models is their ability to handle inputs of varying sizes by sharing the same parameters across all elements. We apply this core idea to the axes of a tensor, proposing a neural network that can process input tensors of varying dimensions using a single set of parameters.

To this end, we introduce a new type of neural network, the Axial Neural Network (XNN), which is equivariant to permutations of a tensor's axes. It achieves this by applying an identical transformation with a shared set of parameters to each axis, thereby treating them as interchangeable elements, similar to the elements of a set or the vertices of a graph. We propose two variants: the set-based XNN and the graph-based XNN.

\subsection{Set-Based Axial Neural Networks}
Set-based Axial Neural Networks (SXNNs) are inspired by Deep Sets~\citep{zaheer2017deep}. They treat the input as the set of all possible axis permutations of a given tensor. Specifically,
\[
&\text{Rank }1:\;\{\bsx\}, & &\text{Rank }3:\;\{\Pi_0(\bsx),\Pi_1(\bsx),\Pi_2(\bsx),\Pi_3(\bsx),\Pi_4(\bsx),\Pi_5(\bsx)\},\\
&\text{Rank }2:\;\{\Pi_0(\bsx),\Pi_1(\bsx)\}, & &\text{Rank }K:\;\{\Pi_0(\bsx),\Pi_1(\bsx),\ldots,\Pi_{K!-1}(\bsx)\},
\] where $\Pi_0(\bsx)=\bsx$.  
A Deep-Set style aggregation as in~\cref{eq:deepset} is then applied:
\[
\label{eq:sxnn}
\bsy=\phi\bigg(\bigoplus_{\Pi\in\bPi}\Pi^{-1}\bigl(\psi(\Pi(\bsx))\bigr)\bigg),
\]
with neural networks $\phi,\psi$, permutation-invariant aggregation $\bigoplus$ (e.g.\ sum, mean, or max), and inverse permutation $\Pi^{-1}$.

\begin{thm}
    \label{thm:sxnn}
    Let $\phi$ be an axis-permutation equivariant function (e.g., another SXNN or pointwise operation). The SXNN in~\cref{eq:sxnn} is axis-permutation equivariant for any rank-$K$ tensor $\bsx \in \bbR^{d_1 \Times d_2 \Times \ldots \Times d_K}$. (\cref{app:sec:sxnn-proof} for proof)
\end{thm}

\textbf{Linear.}
For instance, a simple SXNN may use $\phi$ as the identity and $\psi$ as a linear layer applied along the last axis, followed by max-pooling over the remaining axes for matching the output size. For a rank-3 tensor $\bsx \in \bbR^{H \Times W \Times D}$, it is formalized by
\[
\label{eq:sxnn-linear}
\bsy = \sum_{i=0}^2 \Pi_i^{-1} \left( \operatorname{Pool}_{1,2} \left( \operatorname{Linear}_3 \left( \Pi_i(\bsx) \right) \right) \right),
\]
where $\operatorname{Linear}_3$ applies a linear layer along the third axis ($D$), $\operatorname{Pool}_{1,2}$ pools over the other two axes ($H$ and $W$), and they treat the remaining axis as batch dimensions. Note that only three out of six permutations are required, since pooling is invariant to permutations across the pooled dimensions, i.e. $\text{Pool}(\bbR^{H\Times W\Times D})=\text{Pool}(\bbR^{W\Times H\Times D})$, which omits the redundant permutations. Note also that pooling is equivalent to the approach commonly used in 3D inflation of 2D convolutional layers~\citep{carreira2017quo,mccabe2024multiple}. The transformation flow of the feature sizes in \cref{eq:sxnn-linear} is summarized as
\[
H \Times W \Times D
\begin{Bmatrix}
 \xrightarrow{\Pi_0} H \Times W \Times \bsD \xrightarrow{\text{Linear}_3} \bsH \Times \bsW \Times d \xrightarrow{\text{Pool}_{1,2}} h \Times w \Times d  \xrightarrow{\Pi_0^{-1}} \\
 \xrightarrow{\Pi_1} D \Times H \Times \bsW \xrightarrow{\text{Linear}_3} \bsD \Times \bsH \Times d \xrightarrow{\text{Pool}_{1,2}} d \Times h \Times w \xrightarrow{\Pi_1^{-1}}\\
 \xrightarrow{\Pi_2} W \Times D \Times \bsH \xrightarrow{\text{Linear}_3} \bsW \Times \bsD \Times h \xrightarrow{\text{Pool}_{1,2}} w \Times d \Times h \xrightarrow{\Pi_2^{-1}}
\end{Bmatrix}
\xrightarrow{\bigoplus} h \Times w \Times d\nonumber,
\] where the bold letters indicate the axes that the operation is applied. This construction generalizes to any rank-$K$ tensor with $(K-1)$-dimensional pooling as
\[
\operatorname{(\textbf{Linear})}\quad\bsy = \sum_{i=0}^{K-1} \Pi_i^{-1} \left( \operatorname{Pool}_{1,\ldots,K-1} \left( \operatorname{Linear}_K (\Pi_i(\bsx)) \right) \right).
\] Instead of downsampling like pooling, we may use upsampling operations (e.g., resize) to expand the tensor size if $\operatorname{Linear}_K$ raises the output size.

\textbf{Convolution and Attention.}
For the convolutional layers, we assume the input tensor has an extra channel dimension $C$, i.e., $\bsx \in \bbR^{H \Times W \Times D \Times C}$. The axial convolution is still applied along the spatial axes $H,W,D$. As in the linear case, we can construct the axial convolution using a 1D convolutional layer and the axial self-attention layer applied over one axis. In attention layers, the output sequence length matches the input, so it is unnecessary to apply a pooling or resizing to align output sizes.
\[
\label{eq:xconv}
\hspace{-20pt}\operatorname{(\textbf{Conv})}\sum_{i=0}^{K-1} \Pi^{-1}_i \left( \operatorname{Pool}_{1,\ldots,K-1} \left( \operatorname{Conv1D}_{K} (\Pi_i(\bsx)) \right) \right),\;\;
\operatorname{(\textbf{Attn})}\sum_{i=0}^{K-1} \Pi^{-1}_i \left( \operatorname{SelfAttn}_{K} (\Pi_i(\bsx)) \right).
\]
Interestingly, the set-based axial attention is already used in a recent PDE foundation model~\citep{mccabe2024multiple} for reducing the computational complexity of the Transformer. Splitting the operation across axes reduces the attention overhead. For instance, self-attention over $\bbR^{H \Times W \Times D}$ requires $O((HWD)^2)$ complexity, whereas axial self-attention reduces this to $O(H^2 + W^2 + D^2)$.

\textbf{Non-linearity.}
$\phi$ and $\psi$ in \cref{eq:sxnn} can be arbitrary neural networks, including those with a single non-linearity such as ReLU or Sigmoid. Therefore, using any type of pointwise operation (including non-linearities) does not violate the axis-permutation equivariance of XNN.

\textbf{Expressivity.}
Although SXNNs provide strong expressibility for the dimension-agnostic architecture, their expressivity is inefficient due to their symmetric structure; i.e., they can universally approximate any dimension-agnostic function but require a relatively large width to achieve this. For instance, consider a patch embedding for the Vision Transformer (ViT)~\citep{dosovitskiy2021image} and we assume the patch embedding uses a convolution layer with kernel size 2, stride size 2, and both input and output channels are scalar-valued. In a conventional convolution, the operation on a $2 \Times 2$ pixel patch (2D) can be described as
\[
\label{eq:sxnn-expressivity}
\text{Conv2D}(\bsx):\quad
\begin{bmatrix}
a & b\\
c & d
\end{bmatrix}
*
\begin{bmatrix}
x_1 & x_2\\
x_3 & x_4
\end{bmatrix}
= ax_1 + bx_2 + cx_3 + dx_4,
\]
where $*$ denotes convolution, the first matrix is the kernel, and the second is a $2 \Times 2$ patch from the input image. On the other hand, the axial convolution in~\cref{eq:xconv} with average pooling results in:
\begin{align}
&\begin{Bmatrix}
\text{Conv1D}(\Pi_0(\bsx)):\;
\begin{bmatrix}
a & b\\
a & b
\end{bmatrix}
*
\begin{bmatrix}
x_1 & x_2\\
x_3 & x_4
\end{bmatrix}
=
\begin{bmatrix}
ax_1 + bx_2 \\
ax_3 + bx_4
\end{bmatrix}
\xrightarrow[]{\operatorname{AvgPool}}
\frac{ax_1 + bx_2 + ax_3 + bx_4}{2}
\\[10pt]
\text{Conv1D}(\Pi_1(\bsx)):\;
\begin{bmatrix}
a & b\\
a & b
\end{bmatrix}
*
\begin{bmatrix}
x_1 & x_3\\
x_2 & x_4
\end{bmatrix}
=
\begin{bmatrix}
ax_1 + bx_3 \\
ax_2 + bx_4
\end{bmatrix}
\xrightarrow[]{\operatorname{AvgPool}}
\frac{ax_1 + ax_2 + bx_3 + bx_4}{2}
\end{Bmatrix}\nonumber
\\[4pt]
&\xrightarrow[]{\Sigma}
ax_1 + \tfrac{a + b}{2}x_2 + \tfrac{a + b}{2}x_3 + bx_4
\boldsymbol{
= 
\begin{bmatrix}
a & \tfrac{a + b}{2} \\
\tfrac{a + b}{2} & b
\end{bmatrix}
*
\begin{bmatrix}
x_1 & x_2 \\
x_3 & x_4
\end{bmatrix},
}
\end{align}
which shows that the axial convolution behaves like a convolution with a symmetric kernel. This symmetric kernel structure limits expressivity efficiency, so increasing the number of output channels is often necessary to mitigate this limitation. To address this issue, we also introduce a different type of XNN called the \emph{graph-based XNN}, which avoids the constraint entirely.

\subsection{Graph-Based Axial Neural Networks}
SXNN produces outputs that are inefficient in terms of expressivity compared to the standard neural networks, due to the simple aggregation $\bigoplus$. To overcome this limitation, we can lift the input into an axes-permutation equivariant space, a strategy widely used in the equivariant neural network literature~\citep{cohen2016group,cohen2017steerable,weiler2019general,finzi2020generalizing}, and aggregate them in the intermediate layers as in GNN.

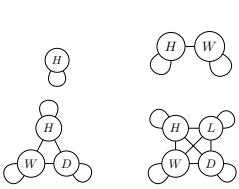
\begin{figure}[t]
\vspace{-30pt}
\noindent
\begin{minipage}[b]{0.25\textwidth}
    \begin{figure}[H]
        \centering
    \hfill
    \resizebox{0.15\textwidth}{!}{%
    \begin{tikzpicture}[scale=.5]
        \node[draw,circle] (H) at (1.5,1.5) {$H$};
        \draw (H);
        \draw (H) to[out=300,in=240,looseness=5] (H);
    \end{tikzpicture}
    }
    \hfill
    \resizebox{0.45\textwidth}{!}{%
    \begin{tikzpicture}[scale=.5]
        \node[draw,circle] (H) at (0,1.5) {$H$};
        \node[draw,circle] (W) at (2,1.5) {$W$};
        \draw (H)--(W);
        \draw (H) to[out=270,in=210,looseness=5] (H);
        \draw (W) to[out=270,in=330,looseness=5] (W);
    \end{tikzpicture}
    }
    \hfill
    \resizebox{0.45\textwidth}{!}{%
    \begin{tikzpicture}[scale=.5]
        \node[draw,circle] (H) at (1,2) {$H$};
        \node[draw,circle] (W) at (0,0) {$W$};
        \node[draw,circle] (D) at (2,0) {$D$};
        \draw (H)--(W) (W)--(D) (D)--(H);
        \draw (H) to[out=60,in=120,looseness=5] (H);
        \draw (W) to[out=180,in=240,looseness=5] (W);
        \draw (D) to[out=0,in=300,looseness=5] (D);
    \end{tikzpicture}
    }
    \hfill
    \resizebox{0.45\textwidth}{!}{%
    \begin{tikzpicture}[scale=.5]
        \node[draw,circle] (H) at (0,2) {$H$};
        \node[draw,circle] (W) at (0,0) {$W$};
        \node[draw,circle] (D) at (2,0) {$D$};
        \node[draw,circle] (L) at (2,2) {$L$};
        \draw (H)--(W) (W)--(D) (D)--(H) (D)--(L) (W)--(L) (H)--(L);
        \draw (H) to[out=180,in=120,looseness=5] (H);
        \draw (W) to[out=180,in=240,looseness=5] (W);
        \draw (D) to[out=0,in=300,looseness=5]   (D);
        \draw (L) to[out=0,in=60,looseness=5]    (L);
    \end{tikzpicture}
    }
        \caption{Axis Graphs.}
        \label{fig:axis-graph}
    \end{figure}
\end{minipage}%
\hfill
\begin{minipage}[b]{0.66\textwidth}
    \begin{figure}[H]
        \centering
        \begin{subfigure}[b]{0.42\textwidth}
        \centering
        \includegraphics[width=\linewidth]{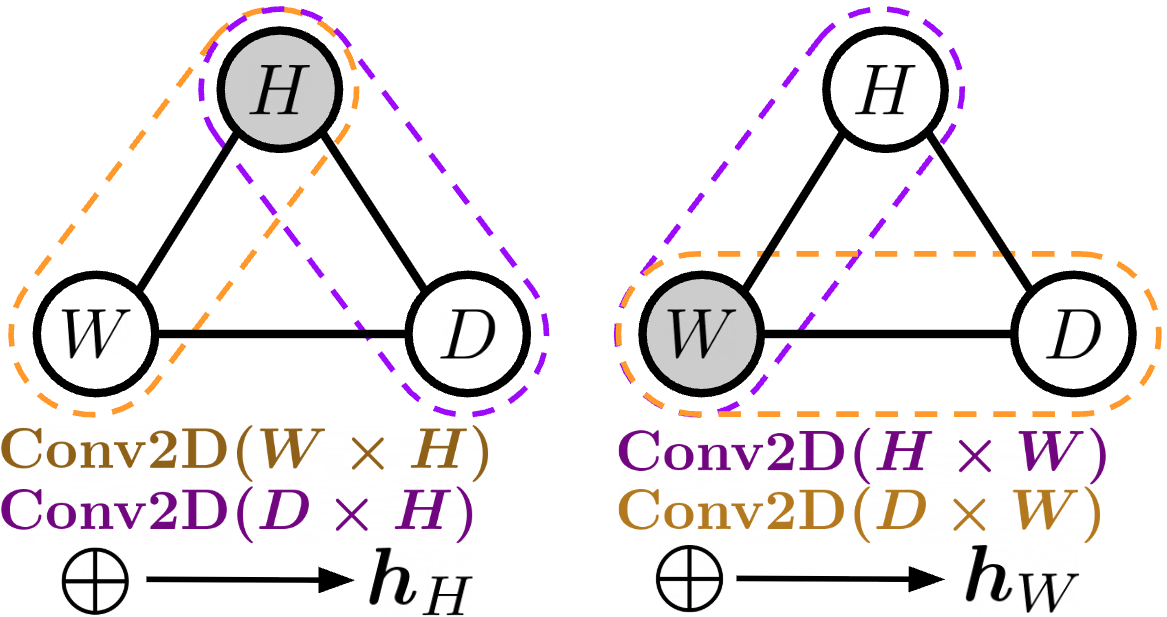}
        \caption{Lifting Layer}
        \label{fig:lifting}
        \end{subfigure}%
        \hfill
        \begin{subfigure}[b]{0.52\textwidth}
        \centering
        \includegraphics[width=\linewidth]{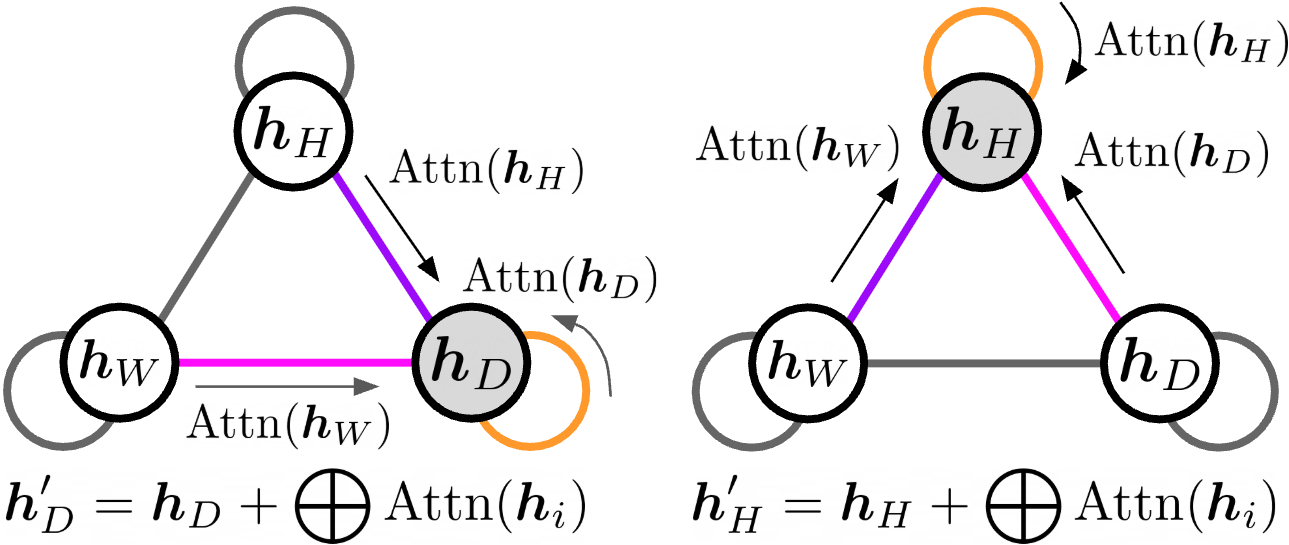}
        \caption{Subsequent Layers}
        \label{fig:subsequent}
        \end{subfigure}
        \caption{Illustration of GXNN in 3D.}
        \label{fig:gxnn}
    \end{figure}
\end{minipage}
\vspace{-10pt}
\end{figure}

\paragraph{Lifting Layer.}
We imagine an undirected graph over the axes such as~\cref{fig:axis-graph}. We now apply the message-passing logic of GNNs as described in~\cref{eq:gnn}. Since the neighbors of $d_1$ are $d_2,\ldots,d_K$ and likewise for the other axes, we can informally express the update rule as
\[
\label{eq:lifting}
\bsh_{d_i} = \phi\bigg(d_i, \bigoplus_{j=1}^K \psi(d_i, d_j) \bigg)
\]
where $\oplus$ denotes an arithmetic form of the permutation-invariant operator $\bigoplus$ and $d_1,d_2,\ldots,d_K$ are used as a conceptual feature representing each axis. The functions $\phi$ and $\psi$ can be chosen based on the architecture or task. Although \cref{eq:lifting} follows parameter-sharing principles and permutation equivariance, in practice, we have to modify it to match the axes order after the aggregation $\oplus$.

For example, in the case of convolutional layers, let $\bsx \in \bbR^{H \Times W \Times D \Times C}$, where $C$ is the number of channels. We can define $\phi$ as the identity function that returns the second argument (i.e. $\phi(A,B)=B$) and $\psi(H, W)$ as a $\operatorname{Conv2D}$ applied over axes $H$ and $W$ with the pooling layer at the end for matching the tensor sizes. We omit $\psi(H,H)$ as it is nontrivial, and the absence of it does not violate the axis permutation equivariance.
When the indices of $\{H,W,D\}$ are $\{1,2,3\}$, \cref{eq:lifting} becomes
\[
\label{eq:lifting-conv}
\bsh_H &= 
\Tr{13}{}\Tr{13}{}^{-1}\operatorname{Pool}_{1}\operatorname{Conv2D}_{2,3}\Tr{13}{}(\bsx)+ \Tr{13}{}\Tr{132}{}^{-1}\operatorname{Pool}_{1}\operatorname{Conv2D}_{2,3}\Tr{132}{}(\bsx),\\
\bsh_W &= 
\Tr{23}{}\Tr{23}{}^{-1}\operatorname{Pool}_{1}\operatorname{Conv2D}_{2,3}\Tr{23}{}(\bsx)+ \Tr{23}{}\Tr{123}{}^{-1}\operatorname{Pool}_{1}\operatorname{Conv2D}_{2,3}\Tr{123}{}(\bsx),\\
\bsh_D &=
\Tr{33}{}\operatorname{Pool}_{1}\operatorname{Conv2D}_{2,3}(\bsx)+\Tr{33}{}\Tr{12}{}^{-1} \operatorname{Pool}_{1}\operatorname{Conv2D}_{2,3}\Tr{12}{}(\bsx),
\] where $\Tr{ijk}{}$ denotes reordering axes $1,2,3$ to $i,j,k$ as explained in~\cref{sec:main:cycle}. Of course, $\phi$ and $\psi$ need not be linear, and it can be a multilayer perceptron. The transformation flow of the feature sizes of $\bsh_H$ in~\cref{eq:lifting-conv} would be:
\[
H \Times W \Times D
\begin{Bmatrix}
\xrightarrow{\Tr{13}{}} D\Times \bsW \Times \bsH \xrightarrow{\text{Conv2D}_{2,3}} \bsD \Times w \Times h \xrightarrow{\text{Pool}_{1}} d \Times w \Times h \xrightarrow{\Tr{13}{}\Tr{13}{}^{-1}}\\
\xrightarrow{\Tr{132}{}} W \Times \bsD \Times \bsH \xrightarrow{\text{Conv2D}_{2,3}} \bsW \Times d \Times h \xrightarrow{\text{Pool}_{1}} w \Times d \Times h \xrightarrow{\Tr{13}{}\Tr{132}{}^{-1}}
\end{Bmatrix}
\xrightarrow{+} d \Times w \Times h=\bsh_{H}.\nonumber
\] Here the channel axis $C$ is omitted for simplicity. A CNN example is also illustrated in~\cref{fig:lifting}.

For 1D and 2D cases, \cref{eq:lifting} reduces to generating the corresponding number of outputs.
\[
\label{eq:lifting-12d}
\text{(1D)}\quad\bsh_H = \phi\left(H, \psi(H,H)\right),\quad\quad
\text{(2D)}\quad\bsh_H &= \phi\left(H, \psi(H,H)\oplus \psi(H,W)\right),\\
\bsh_W &= \phi\left(W, \psi(W,W)\oplus \psi(W,H)\right).
\] The \textbf{1D case} does not consider interaction with the other nodes and $\psi(H,H)$ is still nontrivial. Thus, instead of determining $\psi(H,H)$, we omit the self-edge term $\psi(H,H)$ or $\psi(W,W)$ but rather augment a 1D tensor to a 2D tensor to fully utilize $\operatorname{Conv2D}$. Possible augmentations include the outer product, repetition, and the diagonal matrix. We adopt repetition, which repeats the 1D tensor along a new axis to match the kernel size of Conv2D, and then averages it after lifting to recover the original 1D tensor.

Generalization of the $\operatorname{\textbf{Lifting}}$ layers, \cref{eq:lifting-conv}, to rank-$K$ tensor is described as
\[
\label{eq:lifting-k}
\bsh_{d_i} = \phi\bigg(\Tr{(i)(K)}{}(\bsx), \bigoplus_{j\neq i}^K \Tr{(i)(K)}{}\Tr{(j)(K-1)(i)(K)}{}^{-1}\psi\left(\Tr{(j)(K-1)(i)(K)}{}(\bsx)\right) \bigg),\quad K>1,
\] where $\Tr{(a)(b)(c)(d)}{}=\Tr{abcd}{}$.
\begin{thm}
    \label{thm:lifting} Under some assumptions,
    the lifting layer of GXNN, \cref{eq:lifting-k}, is axis-permutation equivariant for any rank-$K$ (except $K=1$) tensor $\bsx\in\bbR^{d_1\Times d_2\Times\ldots\Times d_K}$. (\cref{app:sec:lifting-proof} for assumptions and proof)
\end{thm}

\begin{figure}[t]
\vspace{-30pt}
    \centering
    \begin{subfigure}[b]{0.36\textwidth}
        \includegraphics[width=\linewidth]{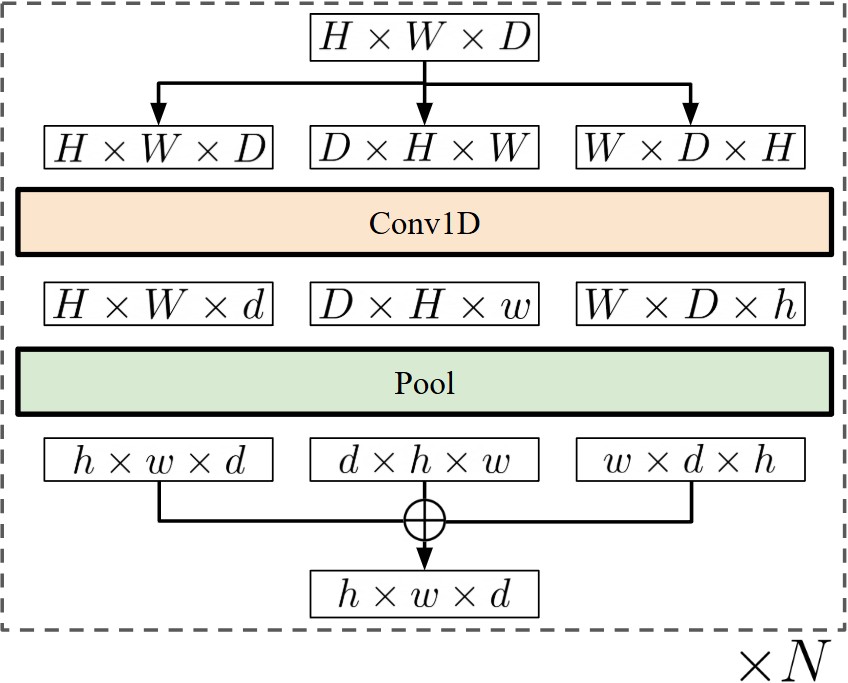}
        \caption{SXCNN}
        \label{fig:image1}
    \end{subfigure}\hfill
    \begin{subfigure}[b]{0.6\textwidth}
        \includegraphics[width=\linewidth]{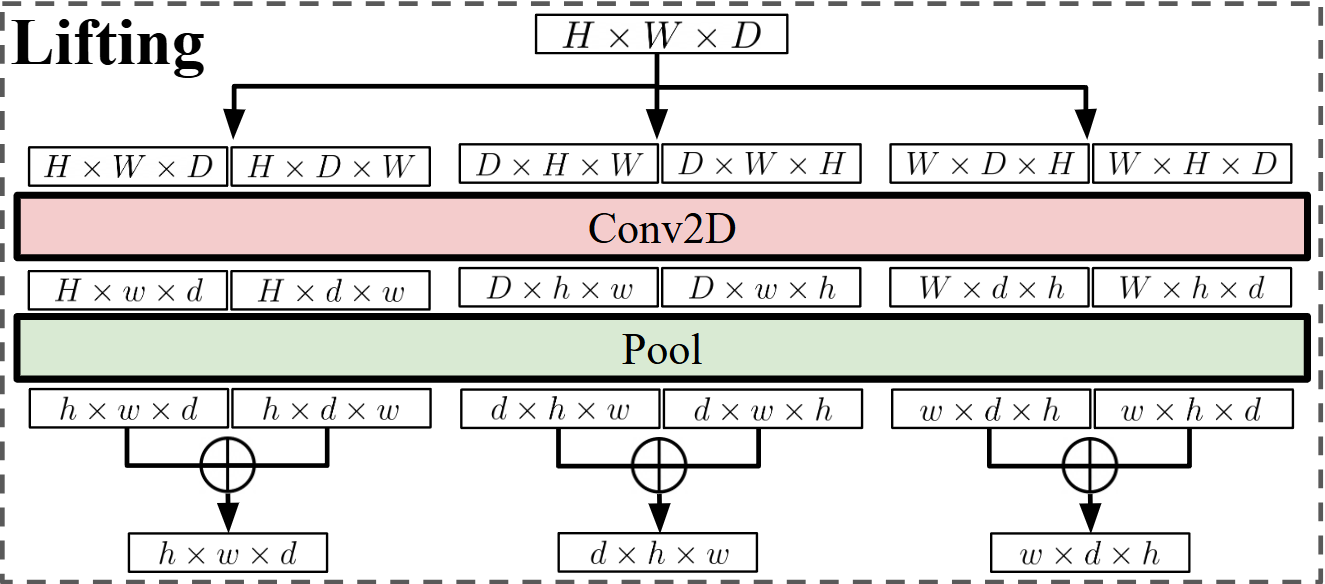}
        \vspace{2mm} 
        \includegraphics[width=\linewidth]{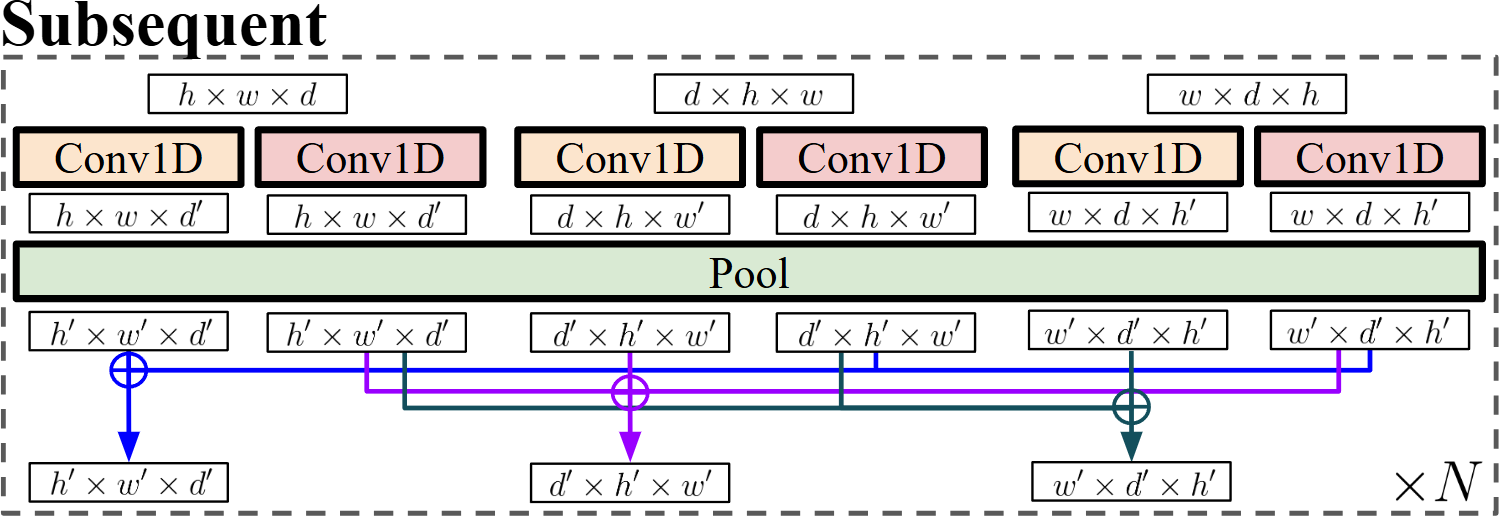}
        \caption{GXCNN}
        \label{fig:image2and3}
    \end{subfigure}
    \caption{An example of SXNN and GXNN for CNN.}
    \label{fig:XCNN}
    \vspace{-10pt}
\end{figure}

\textbf{Subsequent Layers.}
Unlike SXNN, which produces a single feature, we obtain three updated features $\bsh_H$, $\bsh_W$, and $\bsh_D$ in GXNN. Due to the lifting construction, these features are equivariant with respect to permutations of the input axes. In other words, the permutation of input axes results in the permutation of output features with rank 3. Therefore, the subsequent layers will be a GNN whose input is a graph with nodes $\bsh_H$, $\bsh_W$, and $\bsh_D$ as the third graph in~\cref{fig:axis-graph}:
\[
\label{eq:subsequent}
\bsh'_{d_i} = \phi\bigg(\bsh_{d_i}, \bigoplus_{j=1}^K \psi(\bsh_{d_i}, \bsh_{d_j}) \bigg).
\]
Therefore, the subsequent layers should also be dimension-agnostic message-passing architectures, with the lifted features $\bsh_H$, $\bsh_W$, and $\bsh_D$ as graph nodes. \cref{eq:subsequent} also needs to be modified to match the indices order in the aggregation.

For example, in the case of self-attention layers, we can define $\phi$ as a residual path, i.e. $\phi(A,B(A)) = A + B(A)$, and $\psi$ as a self-attention applied over the last axis of the second tensor, i.e. $\psi(A,B)=\operatorname{SelfAttn}_{K}(B)$. Then, as also described in~\cref{fig:subsequent}, the GNN in~\cref{eq:subsequent} becomes
\begin{flalign}
\label{eq:subsequent-attn}
&\bsh'_H = \bsh_H + \Tr{13}{}\Tr{13}{}^{-1}\operatorname{SelfAttn}_3(\bsh_H) +\Tr{13}{}\Tr{23}{}^{-1}\operatorname{SelfAttn}_3(\bsh_W) +\Tr{13}{}\Tr{33}{}^{-1}\operatorname{SelfAttn}_3(\bsh_D),\nonumber\\
&\bsh'_W = \bsh_W + \Tr{23}{}\Tr{13}{}^{-1}\operatorname{SelfAttn}_3(\bsh_H) +\Tr{23}{}\Tr{23}{}^{-1}\operatorname{SelfAttn}_3(\bsh_W) +\Tr{23}{}\Tr{33}{}^{-1}\operatorname{SelfAttn}_3(\bsh_D),\nonumber\\
&\bsh'_D = \bsh_D + \Tr{33}{}\Tr{13}{}^{-1}\operatorname{SelfAttn}_3(\bsh_H) +\Tr{33}{}\Tr{23}{}^{-1}\operatorname{SelfAttn}_3(\bsh_W) + \Tr{33}{}\Tr{33}{}^{-1}\operatorname{SelfAttn}_3(\bsh_D),
\end{flalign} equivalent to
\[
\label{eq:subsequent-attn2}
\lefteqn{\bsm = \Tr{13}{}^{-1}\operatorname{SelfAttn}_3(\bsh_H) +\Tr{23}{}^{-1}\operatorname{SelfAttn}_3(\bsh_W) +\Tr{33}{}^{-1}\operatorname{SelfAttn}_3(\bsh_D),}\\
& \bsh'_H = \bsh_H + \Tr{13}{}(\bsm),\quad \bsh'_W = \bsh_W + \Tr{23}{}(\bsm),\quad \bsh'_D = \bsh_D + \Tr{33}{}(\bsm),
\] where $\Tr{}{}$s are used for aligning the axis order to match the axes in the aggregation. Similarly, the $\operatorname{\textbf{Subsequent}}$ layers in~\cref{eq:subsequent-attn} for a rank-$K$ tensor can be written as
\[
\label{eq:subsequent-k}
\bsh'_{d_i} = \phi\bigg(\bsh_{d_i}, \Tr{(i)(K)}{}\bigoplus_{j=1}^K \Tr{(j)(K)}{}^{-1}\psi(\bsh_{d_j}) \bigg).
\]
\begin{thm}
    \label{thm:subsequent}
    Under some assumptions, the subsequent layers of GXNN in \cref{eq:subsequent-k} are axis-permutation equivariant for any rank-$K$ tensor $\bsh_{d_i}\in\bbR^{d_1\Times d_2\Times\ldots\Times d_K}$. (\cref{app:sec:subsequent-proof} for assumptions and proof)
\end{thm}
\textbf{Pooling Layer.}  
In typical CNNs such as ResNet~\citep{he2016deep}, before computing the output logit values using the linear head, global average pooling or global max pooling is applied over the height and width of the features to aggregate spatial information. Likewise, in GXNN, we obtain $K$ feature tensors through the lifting layer and subsequent layers, and we need to aggregate these features to merge information across axis permutations. Here is an example and its generalized form:
\[
\label{eq:pooling}
(\textbf{rank }3)\quad\bsh' &= \Tr{13}{}^{-1}(\bsh_H') \oplus \Tr{23}{}^{-1}(\bsh_W') \oplus \Tr{33}{}^{-1}(\bsh_D'), & (\textbf{rank }K)\quad\bsh' & = \bigoplus_{i=1}^K\Tr{(i)(K)}{}^{-1}\bsh'_{d_i}.
\] The difference between SXNN and GXNN in a simple CNN architecture can be seen in~\cref{fig:XCNN}. SXCNN naturally satisfies the permutation equivariant structure by repeatedly stacking the same layer. On the other hand, GXCNN requires a lifting layer at the beginning of the network and a pooling layer at the end. In the middle, subsequent layers can be stacked repeatedly. Those XCNNs with added nonlinearity and normalization layers are used in~\cref{sec:main:toy}.

\subsection{Example: Dimension-Agnostic PDE Solver}

One of the important applications is to solve PDEs. Solving PDEs with AI for reducing the cost of numerical PDE solvers, which often requires supercomputers, is a rapidly rising field in modern machine learning~\citep{alkin2024universal,zhou2024unisolver,song2024fmint,liu2024prose}. Each PDE has different spatial dimensionality, and its solutions are represented as a tensor whose elements are points in spatial and temporal grids. Although the neural operator~\citep{kovachki2023neural} has a crucial benefit in solving PDEs, ViT is still commonly used in PDE foundation models due to its strong generalization. The model for Multiple Physics Pretraining (MPP)~\citep{mccabe2024multiple} is one such model that serves powerful performance in multiple 2D PDE training. 

We provide an example of a dimension-agnostic PDE solver by merging GXNN and MPP. MPP consists of patch embedding, multiple attention layers, and patch de-embedding. The patch embedding and patch de-embedding are CNNs. Thus, in the axial implementation, we use the patch embedding as the lifting layer and the rest as the subsequent layers. The attention layers are the same as the $\operatorname{SelfAttn}$ example described in~\cref{eq:subsequent-attn}, and the patch de-embedding is a convolution variant of it. The details and illustrations of the example can be referred to in~\cref{app:sec:pde_solver}.

\section{Related Work}
\label{sec:main:relatedwork}

Several studies have explored dimension-agnostic architectures.
\citet{levin2024any} proposed any-dimensional equivariant neural networks that leverage representation stability from algebraic topology, enabling models trained on fixed input dimensions to generalize to arbitrary sizes.
Similarly, \citet{lee2025dimension} introduced dimension-agnostic neural processes, which incorporate a dimension aggregator block to unify inputs of varying dimensions into a shared latent space. These approaches offer strong potential for constructing flexible operators that scale beyond conventional grid-dependent solvers.
From a practical standpoint, multi-modal training, particularly joint training on images and videos, has also been widely explored~\citep{dandi2020jointly,xu2023mplug2,girdhar2022omnivore,li2023uniformerv2,chen2024gentron}. However, these methods typically treat images as individual video frames and either introduce temporal attention layers or use separate embedders for video, without adopting a truly dimension-agnostic approach.

\section{Experiments}
\label{sec:main:experiment}

\subsection{Toy Dataset: Gaussian Process Kernel Prediction}
\label{sec:main:toy}
\begin{wrapfigure}{l}{0.35\textwidth}
\vspace{-15pt}
    \centering
    \includegraphics[width=0.35\textwidth]{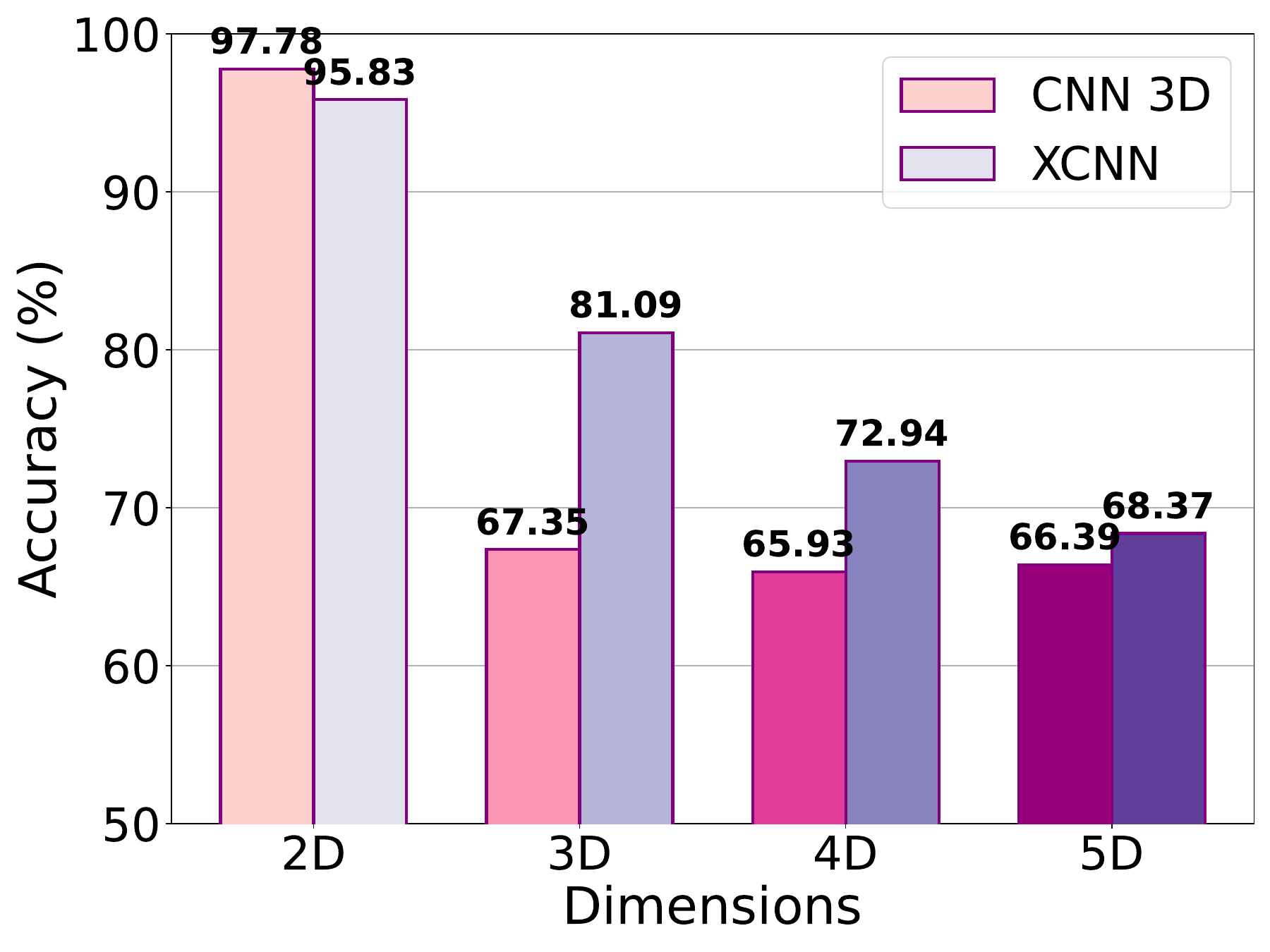}
    \caption{Accuracy on GP kernel prediction.}
    \label{fig:toy}
\vspace{-10pt}
\end{wrapfigure}
To demonstrate the dimension-free training and inference capabilities of XNN, we construct a toy dataset of synthetic Gaussian process (GP) data. The task is binary classification between two kernels: radial basis function (RBF) and periodic, given multi-dimensional data (2D, 3D, 4D, and 5D) randomly sampled from the synthetic GP. We build two model to train it: a CNN with 3D convolutions and an XCNN with graph-based axial convolutions composed of 2D convolutions.
Although our architecture naturally supports multi-dimensional data, the 3D CNN is restricted to only 3D spatial inputs. Therefore, for 2D data, we pad with zeros to construct 3D-like inputs, and for 4D and 5D data, we flatten the last axes to obtain 3D shapes (e.g., 5D: \(16 \Times 16 \Times 16 \Times 16 \Times 16 \Times C \to 16 \Times 16 \Times 4096 \Times C\), where \(C\) is the channel dimension). Further experimental details are provided in~\cref{app:sec:detail}.
As shown in~\cref{fig:toy}, XCNN built with 2D convolutions exhibits superior test accuracy across all dimensions. 
Notably, despite using 3D convolutions, CNN-3D performs poorly on higher-dimensional data, including 3D, highlighting the robustness of XNN’s dimension-agnostic design.

\textbf{Comparison Between SXNN and GXNN.}
SXNN exhibits a trade-off between computational efficiency and expressivity inefficiency as described in~\cref{eq:sxnn-expressivity}. To examine this and verify the necessity of GXNN, we build a set-based axial CNN (SXCNN) composed of Conv1Ds and a graph-based axial CNN (GXCNN) from the previous experiment. We then compare their number of parameters (\#Params), wall clock time (W. Clock) of forward computation, and inference performance (Acc.) using the GP toy dataset. \cref{tab:sxnn_gxnn} shows the comparisons, where SXCNN-L is an enlarged version of SXCNN for fair comparison with GXCNN. Note that SXCNN with the same depth and width as GXCNN has $6\times$ fewer parameters and 80\% reduced wall clock time, but its performance degrades in high dimensions. After increasing the depth and width, it performs fairly well in high dimensions, but GXCNN still exhibits superior performance, indicating the necessity of GXNN.

\subsection{PDE Solver Foundation Models}

In this experiment, we evaluate the effectiveness of XNN as an architecture for multi-dimensional training. A compelling use case would be multi-PDE solution training, which involves different PDEs with varying dimensionalities. We train PDE solver foundation models on the solutions of time-homogeneous PDEs drawn from a variety of physical systems. The data are sourced from widely used PDE solution benchmark datasets: PDEBench~\citep{takamoto2022pdebench} and PDEArena~\citep{gupta2022towards}. The list of PDEs and their details are provided in~\cref{app:sec:pdes}.

We implemented our architecture in two state-of-the-art PDE foundation model baselines: CViT~\citep{wang2024cvit} and MPP~\citep{mccabe2024multiple}. Both are based on ViT~\citep{dosovitskiy2021image}, incorporating patch embedding and multiple attention layers with some revisions optimized for PDE learning. Note that these baselines are designed only for 2D data, as the 2D convolutional layer for patch embedding is dimension-dependent, though the attention layers are not. We build their XNN variants, termed X-CViT and X-MPP, which include axial linears, axial convolutions, and axial attentions. The specifications are detailed in~\cref{app:sec:variant}.

Throughout the experiments, we followed the basic training procedures of baselines such as CViT and MPP, which take a few timesteps as input and predict the next timestep of a PDE solution. We set CViT, X-CViT to take $s=2$ timesteps as input, and MPP, X-MPP to take $s=4$, in contrast to the original MPP paper, which used $s=16$. We also replace InstanceNorm~\citep{ulyanov2016instance} of MPP by LayerNorm~\citep{ba2016layer} due to training instability observed on 1D and 3D data. The evaluation metric is the Normalized Root Mean Squared Error (NRMSE), defined in~\cref{app:sec:detail}.

The handling of domain-specific features such as boundary conditions, geometry, and time is determined by the baseline methods we modified for XNN; i.e., X-MPP follows the handling method of MPP. Additionally, the baseline methods target PDE solutions defined only on regular grids. According to MPP, the boundary conditions, geometry, and time are not separately input to the model. Instead, MPP is pretrained on multiple PDE solutions with varying boundary conditions and equations, allowing it to learn general patterns of PDE solutions. Since we eventually finetune the model on a specific PDE with known boundary conditions and geometry, it is not necessary to encode them separately as inputs. The only thing MPP handles separately is the periodicity of the boundary condition. Depending on the periodicity of the boundary condition, MPP determines whether to use sequential position bias or periodic position bias in the attention layer.

To isolate the effect of our architectural modifications, we use the smallest versions of CViT and MPP as backbones: CViT-S and MPP-Ti, with 12M and 7M parameters, respectively. Unlike \citet{mccabe2024multiple}, we exclude incompressible fluid dynamics from training due to its large data size relative to model capacity. The model size is smaller than the pretraining dataset ($\sim$40M), so pretraining offers limited benefit over training from scratch, but it is sufficient to highlight the advantage of multidimensional training.

\begin{figure}[t]
\vspace{-20pt}
\centering
\begin{minipage}[t]{0.32\textwidth}
\centering
\captionof{table}{SXCNN vs. GXCNN.}
\label{tab:sxnn_gxnn}
\resizebox{\linewidth}{!}{%
\begin{tabular}{cccc}
\toprule
 & SXCNN & SXCNN-L & GXCNN \\
\midrule
Depth & 4 & 5 &  4\\
Width & 128 & 256 & 128 \\
\#Params & 150K & 791K & 899K \\
W. Clock & 80ms & 98ms & 101ms \\
2D Acc. & 95.94 & 92.60 & 95.45 \\
3D Acc. & 63.68 & 85.72 & 79.85 \\
4D Acc. & 42.54 & 55.18 & 70.86 \\
5D Acc. & 62.09 & 62.64 & 70.48 \\
\bottomrule
\end{tabular}%
}

\end{minipage}
\hfill
\begin{minipage}[t]{0.65\textwidth}
\centering
\captionof{table}{Test NRMSE of PDE solvers on 2D PDEs.}
\label{tab:expressivity}
\resizebox{\linewidth}{!}{%
\begin{tabular}{cccccccccc}
\toprule
Model & Pretrain & FineTune & DR & NS & SWE & CFD M0.1 & CFD M1.0 \\
 \midrule
CViT & $\times$ & $\times$ & 0.0389 & \UL{0.1078}\textsuperscript{\textdagger} & \UL{0.1876}\textsuperscript{\textdagger} & - & -  \\
X-CViT & $\times$ & $\times$ & \UL{0.0382} & 0.1148\textsuperscript{\textdagger} & 0.1948\textsuperscript{\textdagger} & - & -  \\
\midrule
MPP & $\times$ & $\times$ & 0.0157 & - & 0.0015 & 0.0132 & 0.0181  \\
X-MPP & $\times$ & $\times$ & \UL{0.0118} & - & \UL{0.0012} & \UL{0.0118} & \UL{0.0163}  \\
\midrule
MPP & 2D & $\times$ & 0.0447 &  & 0.0087 & \UL{0.0404}$^*$ & \UL{0.0499}$^*$  \\
MPP & 1D,2D & $\times$ & 0.2183 & - & 0.0265 & 0.1881$^*$ & 0.2199$^*$  \\
X-MPP & 1D,2D,3D & $\times$ & \UL{0.0430} & - & \UL{0.0086} & 0.0428$^*$ & 0.0517$^*$ \\
\midrule
MPP & 2D & \checkmark & 0.0516  &  - & 0.0022 & 0.0319$^*$ & 0.0422$^*$ \\
X-MPP & 1D,2D,3D & \checkmark & \UL{0.0058}  & - & \UL{0.0011}  & \UL{0.0209}$^*$  & \UL{0.0294}$^*$  \\
\bottomrule
\end{tabular}%
}

\end{minipage}
\vspace{-10pt}
\end{figure}

\textbf{Expressivity of XNN.}
We evaluated the architectural expressivity of XNN by measuring how well it performs compared to dimension-specific models in three different settings: single PDE training, multiple PDE training, and single PDE finetuning. Since the baselines are designed for 2D PDEs, we evaluate performance only on 2D PDEs: Diffusion-Reaction (DR), Incompressible Navier-Stokes (NS), Shallow Water Equation (SWE), and Compressible Fluid Dynamics (CFD) with Mach numbers $M=0.1$ and $1.0$. We measure NRMSE for each PDE. Notably, for 1D-2D joint training in MPP, we convert 1D PDE solutions to 2D by padding with zero values.

The results are shown in~\cref{tab:expressivity}. The notation (\textdagger)~in the table denotes evaluation on PDEArena, meaning the rest are on PDEBench. ($^*$) indicates results trained with both CFD M0.1 and CFD M1.0, meaning more challenging setup compared to training respectively. The \UL{underscores} denote the best result compared to the competitors. Since we report based on the baseline codebase, the empty slots denote PDEs that the codebase does not support. In every scenario, the axial variants exhibit competitive results compared to the non-axial baselines. In particular, due to the benefit of multidimensional pretraining, the finetuned X-MPP (trained with 1D, 2D, and 3D data) outperforms MPP trained only with 2D. Note that pretraining MPP with both 1D and 2D leads to significant degradation, as it fails to learn a unified representation space across dimensions.

\begin{figure}[t]
\vspace{-20pt}
    \centering
    \includegraphics[width=\textwidth]{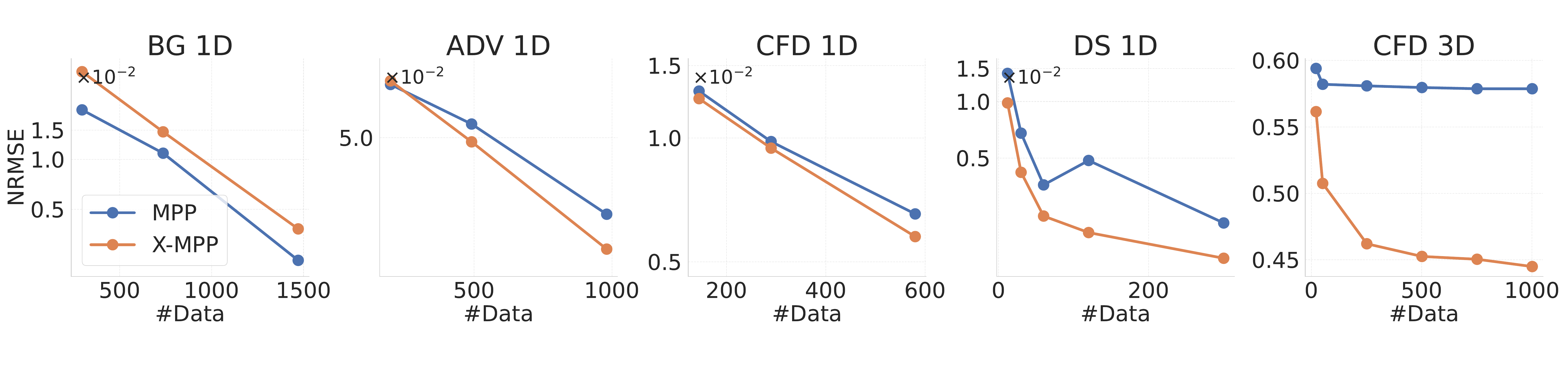}
    \vspace{-25pt}
    \caption{Test NRMSE of finetuning on unseen dimensions (1D and 3D).}
    \label{fig:unseen}
    \vspace{-12pt}
\end{figure}

\textbf{Unseen Dimension Generalization.}
In this experiment, we demonstrate the dimension-generalization (few-shot learning) ability of XNN on data from unseen dimensions. To do so, we pretrain both MPP and X-MPP on 2D PDE data and compare their finetuning performance on unseen 1D and 3D PDEs, demonstrating the benefit of multi-dimensional pretraining in XNN-based foundation models. We use three 1D PDEs and one 3D PDE: Diffusion Sorption (DS), Burgers' equation (BG), 1D Compressible Fluid Dynamics (CFD 1D), and 3D Compressible Fluid Dynamics (CFD 3D). We evaluate using the same metric, NRMSE, and compare how finetuning performance improves as the size of the given dataset increases.

For 3D finetuning of MPP, we use the inflation technique described in~\citet{mccabe2024multiple}. It repeats a $P \times P$ kernel of the 2D convolution layers $P$ times and divides by $P$. The weights of the linear projection for the additional variable in the 3D PDE are initialized with the average of the trained weights corresponding to the existing variables. For 1D finetuning of MPP, we augment the input tensor from 1D to 2D only in the patch embedding at the beginning and the patch de-embedding at the end, so that the attention layers in the middle operate on purely 1D PDEs. In contrast, our model naturally extends to 3D and reduces to 1D. 
For X-MPP, we use a shared linear projection across different PDEs for unified representation learning. This contrasts with MPP, where a different linear projection learns each PDE during fine-tuning.

As shown in~\cref{fig:unseen}, except for BG 1D, X-MPP (orange) accelerates the finetuning performance with only a small amount of data. In particular for 3D, the redundancy of the inflated layers in MPP (blue) limits 3D generalization, whereas the dimension-agnostic architecture is free from this issue, thereby efficiently utilizing the representation trained from 2D PDEs.
\section{Conclusion}
\label{sec:main:conclusion}

In this work, we introduce XNNs, a family of efficient and expressive architectures designed to be agnostic to input dimensionality. 
Motivated by the limitations of prior PDE models constrained to fixed dimensions, XNNs leverage axis permutation equivariance over tensor axes to naturally generalize across 1D, 2D, and 3D domains, which can be applied to develop a dimension-agnostic PDE solver foundation model.

In the PDE solving problem, our empirical results highlight the benefits of multidimensional pretraining and the superior finetuning ability of XNNs compared to the dimension-specific models such as MPP.
These findings underscore the importance of designing foundation models capable of operating across varying spatial dimensions, which is a critical step toward scalable and adaptable scientific machine learning systems.

\textbf{Limitations and Future Work.}
An axial version of the cross attention between two tensors of different dimensions is not yet fully explored. Addressing this issue as a natural extension of this work is a valuable future direction that would further widen the usability of XNNs. 
Additionally, improving the computational efficiency of GXNN, especially for high-dimensional data, is also a promising direction towards the practical deployment of XNNs in large-scale scientific simulations.

\newpage
\section*{Acknowledgement}
We thank \emph{Heejun Lee} for his thoughtful technical support, \emph{Jules Berman} for his insightful concerns, and \emph{Kyunghyun Cho} for facilitating our wonderful collaborative research.
This work was partly supported by Institute of Information \& communications Technology Planning \& Evaluation(IITP) grant funded by the Korea government(MSIT) (No.RS-2019-II190075, Artificial Intelligence Graduate School Program(KAIST); No.RS-2024-00509279, Global AI Frontier Lab; No.RS-2022-II220713, Meta-learning Applicable to Real-world Problems) and Artificial intelligence industrial convergence cluster development project funded by the Ministry of Science and ICT(MSIT, Korea) \& Gwangju Metropolitan City. H. Yang was supported by the National Research Foundation of Korea(NRF) grant funded by the Korean Government(MSIT) (No. RS-2023-00279680).
\bibliographystyle{plainnat}
\bibliography{references}

\begin{thebibliography}{41}
\providecommand{\natexlab}[1]{#1}
\providecommand{\url}[1]{\texttt{#1}}
\expandafter\ifx\csname urlstyle\endcsname\relax
  \providecommand{\doi}[1]{doi: #1}\else
  \providecommand{\doi}{doi: \begingroup \urlstyle{rm}\Url}\fi

\bibitem[Alkin et~al.(2024)Alkin, F{\"{u}}rst, Schmid, Gruber, Holzleitner, and Brandstetter]{alkin2024universal}
Benedikt Alkin, Andreas F{\"{u}}rst, Simon Schmid, Lukas Gruber, Markus Holzleitner, and Johannes Brandstetter.
\newblock Universal physics transformers: {A} framework for efficiently scaling neural operators.
\newblock In \emph{Advances in Neural Information Processing Systems 38: Annual Conference on Neural Information Processing Systems 2024, NeurIPS 2024, Vancouver, BC, Canada, December 10 - 15, 2024}, 2024.

\bibitem[Ba et~al.(2016)Ba, Kiros, and Hinton]{ba2016layer}
Lei~Jimmy Ba, Jamie~Ryan Kiros, and Geoffrey~E. Hinton.
\newblock Layer normalization.
\newblock \emph{CoRR}, abs/1607.06450, 2016.
\newblock URL \url{http://arxiv.org/abs/1607.06450}.

\bibitem[Bloem{-}Reddy and Teh(2020)]{bloem-reddy2020probabilistic}
Benjamin Bloem{-}Reddy and Yee~Whye Teh.
\newblock Probabilistic symmetries and invariant neural networks.
\newblock \emph{J. Mach. Learn. Res.}, 21:\penalty0 90:1--90:61, 2020.
\newblock URL \url{https://jmlr.org/papers/v21/19-322.html}.

\bibitem[Bordelon et~al.(2024)Bordelon, Atanasov, and Pehlevan]{bordelon2024dynamical}
Blake Bordelon, Alexander~B. Atanasov, and Cengiz Pehlevan.
\newblock A dynamical model of neural scaling laws.
\newblock In \emph{Forty-first International Conference on Machine Learning, {ICML} 2024, Vienna, Austria, July 21-27, 2024}. OpenReview.net, 2024.

\bibitem[Bradbury et~al.(2018)Bradbury, Frostig, Hawkins, Johnson, Leary, Maclaurin, Necula, Paszke, Vander{P}las, Wanderman-{M}ilne, and Zhang]{bradbury2018jax}
James Bradbury, Roy Frostig, Peter Hawkins, Matthew~James Johnson, Chris Leary, Dougal Maclaurin, George Necula, Adam Paszke, Jake Vander{P}las, Skye Wanderman-{M}ilne, and Qiao Zhang.
\newblock {JAX}: composable transformations of {P}ython+{N}um{P}y programs, 2018.
\newblock URL \url{http://github.com/jax-ml/jax}.

\bibitem[Carreira and Zisserman(2017)]{carreira2017quo}
Jo{\~{a}}o Carreira and Andrew Zisserman.
\newblock Quo vadis, action recognition? {A} new model and the kinetics dataset.
\newblock In \emph{2017 {IEEE} Conference on Computer Vision and Pattern Recognition, {CVPR} 2017, Honolulu, HI, USA, July 21-26, 2017}, pages 4724--4733. {IEEE} Computer Society, 2017.

\bibitem[Chen et~al.(2024)Chen, Xu, Ren, Cong, He, Xie, Sinha, Luo, Xiang, and P{\'{e}}rez{-}R{\'{u}}a]{chen2024gentron}
Shoufa Chen, Mengmeng Xu, Jiawei Ren, Yuren Cong, Sen He, Yanping Xie, Animesh Sinha, Ping Luo, Tao Xiang, and Juan{-}Manuel P{\'{e}}rez{-}R{\'{u}}a.
\newblock Gentron: Diffusion transformers for image and video generation.
\newblock In \emph{{IEEE/CVF} Conference on Computer Vision and Pattern Recognition, {CVPR} 2024, Seattle, WA, USA, June 16-22, 2024}, pages 6441--6451. {IEEE}, 2024.
\newblock \doi{10.1109/CVPR52733.2024.00616}.
\newblock URL \url{https://doi.org/10.1109/CVPR52733.2024.00616}.

\bibitem[Cohen and Welling(2016)]{cohen2016group}
Taco Cohen and Max Welling.
\newblock Group equivariant convolutional networks.
\newblock In \emph{Proceedings of The 33rd International Conference on Machine Learning (ICML 2016)}, 2016.

\bibitem[Cohen and Welling(2017)]{cohen2017steerable}
Taco~S. Cohen and Max Welling.
\newblock Steerable cnns.
\newblock In \emph{International Conference on Learning Representations (ICLR)}, 2017.

\bibitem[Dandi et~al.(2020)Dandi, Das, Singhal, Namboodiri, and Rai]{dandi2020jointly}
Yatin Dandi, Aniket Das, Soumye Singhal, Vinay~P. Namboodiri, and Piyush Rai.
\newblock Jointly trained image and video generation using residual vectors.
\newblock In \emph{{IEEE} Winter Conference on Applications of Computer Vision, {WACV} 2020, Snowmass Village, CO, USA, March 1-5, 2020}, pages 3017--3031. {IEEE}, 2020.
\newblock \doi{10.1109/WACV45572.2020.9093308}.
\newblock URL \url{https://doi.org/10.1109/WACV45572.2020.9093308}.

\bibitem[de~Finetti and B.(1931)]{finetti1931funzione}
de~Finetti and B.
\newblock Funzione caratteristica di un fenomeno aleatorio.
\newblock \emph{Attidella {R}. {Academia} {Nazionale} dei {Lincei}, {Serie}}, 6.\penalty0 (4):\penalty0 251299., 1931.

\bibitem[Dosovitskiy et~al.(2021)Dosovitskiy, Beyer, Kolesnikov, Weissenborn, Zhai, Unterthiner, Dehghani, Minderer, Heigold, Gelly, Uszkoreit, and Houlsby]{dosovitskiy2021image}
Alexey Dosovitskiy, Lucas Beyer, Alexander Kolesnikov, Dirk Weissenborn, Xiaohua Zhai, Thomas Unterthiner, Mostafa Dehghani, Matthias Minderer, Georg Heigold, Sylvain Gelly, Jakob Uszkoreit, and Neil Houlsby.
\newblock An image is worth 16x16 words: Transformers for image recognition at scale.
\newblock In \emph{9th International Conference on Learning Representations, {ICLR} 2021, Virtual Event, Austria, May 3-7, 2021}. OpenReview.net, 2021.

\bibitem[Finzi et~al.(2020)Finzi, Stanton, Izmailov, and Wilson]{finzi2020generalizing}
Marc Finzi, Samuel Stanton, Pavel Izmailov, and Andrew~Gordon Wilson.
\newblock Generalizing convolutional neural networks for equivariance to lie groups on arbitrary continuous data.
\newblock In \emph{Proceedings of The 37th International Conference on Machine Learning (ICML 2020)}, 2020.

\bibitem[Girdhar et~al.(2022)Girdhar, Singh, Ravi, van~der Maaten, Joulin, and Misra]{girdhar2022omnivore}
Rohit Girdhar, Mannat Singh, Nikhila Ravi, Laurens van~der Maaten, Armand Joulin, and Ishan Misra.
\newblock Omnivore: {A} single model for many visual modalities.
\newblock In \emph{{IEEE/CVF} Conference on Computer Vision and Pattern Recognition, {CVPR} 2022, New Orleans, LA, USA, June 18-24, 2022}, pages 16081--16091. {IEEE}, 2022.
\newblock \doi{10.1109/CVPR52688.2022.01563}.
\newblock URL \url{https://doi.org/10.1109/CVPR52688.2022.01563}.

\bibitem[Gori et~al.(2005)Gori, Monfardini, and Scarselli]{gori2005new}
Marco Gori, Gabriele Monfardini, and Franco Scarselli.
\newblock A new model for learning in graph domains.
\newblock In \emph{{IEEE} International Joint Conference on Neural Networks, {IJCNN} 2005, Montreal, QC, Canada, July 31 - August 4, 2005}, pages 729--734. {IEEE}, 2005.

\bibitem[Gupta and Brandstetter(2022)]{gupta2022towards}
Jayesh~K Gupta and Johannes Brandstetter.
\newblock Towards multi-spatiotemporal-scale generalized pde modeling.
\newblock \emph{arXiv preprint arXiv:2209.15616}, 2022.

\bibitem[Hao et~al.(2024)Hao, Su, Liu, Berner, Ying, Su, Anandkumar, Song, and Zhu]{hao2024dpot}
Zhongkai Hao, Chang Su, Songming Liu, Julius Berner, Chengyang Ying, Hang Su, Anima Anandkumar, Jian Song, and Jun Zhu.
\newblock {DPOT:} auto-regressive denoising operator transformer for large-scale {PDE} pre-training.
\newblock In \emph{Forty-first International Conference on Machine Learning, {ICML} 2024, Vienna, Austria, July 21-27, 2024}. OpenReview.net, 2024.

\bibitem[Havrilla and Liao(2024)]{havrilla2024understanding}
Alexander Havrilla and Wenjing Liao.
\newblock Understanding scaling laws with statistical and approximation theory for transformer neural networks on intrinsically low-dimensional data.
\newblock In \emph{Advances in Neural Information Processing Systems 38: Annual Conference on Neural Information Processing Systems 2024, NeurIPS 2024, Vancouver, BC, Canada, December 10 - 15, 2024}, 2024.

\bibitem[He et~al.(2016)He, Zhang, Ren, and Sun]{he2016deep}
Kaiming He, Xiangyu Zhang, Shaoqing Ren, and Jian Sun.
\newblock Deep residual learning for image recognition.
\newblock In \emph{2016 {IEEE} Conference on Computer Vision and Pattern Recognition, {CVPR} 2016, Las Vegas, NV, USA, June 27-30, 2016}, pages 770--778. {IEEE} Computer Society, 2016.

\bibitem[Heek et~al.(2024)Heek, Levskaya, Oliver, Ritter, Rondepierre, Steiner, and van {Z}ee]{heek2024flax}
Jonathan Heek, Anselm Levskaya, Avital Oliver, Marvin Ritter, Bertrand Rondepierre, Andreas Steiner, and Marc van {Z}ee.
\newblock {F}lax: A neural network library and ecosystem for {JAX}, 2024.
\newblock URL \url{http://github.com/google/flax}.

\bibitem[Herde et~al.(2024)Herde, Raonic, Rohner, K{\"{a}}ppeli, Molinaro, de~B{\'{e}}zenac, and Mishra]{herde2024poseidon}
Maximilian Herde, Bogdan Raonic, Tobias Rohner, Roger K{\"{a}}ppeli, Roberto Molinaro, Emmanuel de~B{\'{e}}zenac, and Siddhartha Mishra.
\newblock Poseidon: Efficient foundation models for pdes.
\newblock In \emph{Advances in Neural Information Processing Systems 38: Annual Conference on Neural Information Processing Systems 2024, NeurIPS 2024, Vancouver, BC, Canada, December 10 - 15, 2024}, 2024.

\bibitem[Kaplan et~al.(2020)Kaplan, McCandlish, Henighan, Brown, Chess, Child, Gray, Radford, Wu, and Amodei]{kaplan2020scaling}
Jared Kaplan, Sam McCandlish, Tom Henighan, Tom~B. Brown, Benjamin Chess, Rewon Child, Scott Gray, Alec Radford, Jeffrey Wu, and Dario Amodei.
\newblock Scaling laws for neural language models, 2020.
\newblock URL \url{https://arxiv.org/abs/2001.08361}.

\bibitem[Kovachki et~al.(2023)Kovachki, Li, Liu, Azizzadenesheli, Bhattacharya, Stuart, and Anandkumar]{kovachki2023neural}
Nikola~B. Kovachki, Zongyi Li, Burigede Liu, Kamyar Azizzadenesheli, Kaushik Bhattacharya, Andrew~M. Stuart, and Anima Anandkumar.
\newblock Neural operator: Learning maps between function spaces with applications to pdes.
\newblock \emph{J. Mach. Learn. Res.}, 24:\penalty0 89:1--89:97, 2023.
\newblock URL \url{https://jmlr.org/papers/v24/21-1524.html}.

\bibitem[Lee et~al.(2025)Lee, Jang, Lee, and Lee]{lee2025dimension}
Hyungi Lee, Chaeyun Jang, Dongbok Lee, and Juho Lee.
\newblock Dimension agnostic neural processes, 2025.
\newblock URL \url{https://arxiv.org/abs/2502.20661}.

\bibitem[Levin and D{\'{\i}}az(2024)]{levin2024any}
Eitan Levin and Mateo D{\'{\i}}az.
\newblock Any-dimensional equivariant neural networks.
\newblock In Sanjoy Dasgupta, Stephan Mandt, and Yingzhen Li, editors, \emph{International Conference on Artificial Intelligence and Statistics, 2-4 May 2024, Palau de Congressos, Valencia, Spain}, volume 238 of \emph{Proceedings of Machine Learning Research}, pages 2773--2781. {PMLR}, 2024.
\newblock URL \url{https://proceedings.mlr.press/v238/levin24a.html}.

\bibitem[Li et~al.(2023)Li, Wang, He, Li, Wang, Wang, and Qiao]{li2023uniformerv2}
Kunchang Li, Yali Wang, Yinan He, Yizhuo Li, Yi~Wang, Limin Wang, and Yu~Qiao.
\newblock Uniformerv2: Unlocking the potential of image vits for video understanding.
\newblock In \emph{{IEEE/CVF} International Conference on Computer Vision, {ICCV} 2023, Paris, France, October 1-6, 2023}, pages 1632--1643. {IEEE}, 2023.
\newblock \doi{10.1109/ICCV51070.2023.00157}.
\newblock URL \url{https://doi.org/10.1109/ICCV51070.2023.00157}.

\bibitem[Liu et~al.(2024)Liu, Sun, He, Pinney, Zhang, and Schaeffer]{liu2024prose}
Yuxuan Liu, Jingmin Sun, Xinjie He, Griffin Pinney, Zecheng Zhang, and Hayden Schaeffer.
\newblock Prose-fd: A multimodal pde foundation model for learning multiple operators for forecasting fluid dynamics, 2024.
\newblock URL \url{https://arxiv.org/abs/2409.09811}.

\bibitem[McCabe et~al.(2024)McCabe, Blancard, Parker, Ohana, Cranmer, Bietti, Eickenberg, Golkar, Krawezik, Lanusse, Pettee, Tesileanu, Cho, and Ho]{mccabe2024multiple}
Michael McCabe, Bruno~R{\'{e}}galdo{-}Saint Blancard, Liam~Holden Parker, Ruben Ohana, Miles~D. Cranmer, Alberto Bietti, Michael Eickenberg, Siavash Golkar, G{\'{e}}raud Krawezik, Fran{\c{c}}ois Lanusse, Mariel Pettee, Tiberiu Tesileanu, Kyunghyun Cho, and Shirley Ho.
\newblock Multiple physics pretraining for spatiotemporal surrogate models.
\newblock In \emph{Advances in Neural Information Processing Systems 38: Annual Conference on Neural Information Processing Systems 2024, NeurIPS 2024, Vancouver, BC, Canada, December 10 - 15, 2024}, 2024.

\bibitem[Paszke et~al.(2019)Paszke, Gross, Massa, Lerer, Bradbury, Chanan, Killeen, Lin, Gimelshein, Antiga, Desmaison, Köpf, Yang, DeVito, Raison, Tejani, Chilamkurthy, Steiner, Fang, Bai, and Chintala]{paszke2019pytorch}
Adam Paszke, Sam Gross, Francisco Massa, Adam Lerer, James Bradbury, Gregory Chanan, Trevor Killeen, Zeming Lin, Natalia Gimelshein, Luca Antiga, Alban Desmaison, Andreas Köpf, Edward Yang, Zach DeVito, Martin Raison, Alykhan Tejani, Sasank Chilamkurthy, Benoit Steiner, Lu~Fang, Junjie Bai, and Soumith Chintala.
\newblock Pytorch: An imperative style, high-performance deep learning library, 2019.
\newblock URL \url{https://arxiv.org/abs/1912.01703}.

\bibitem[Scarselli et~al.(2009)Scarselli, Gori, Tsoi, Hagenbuchner, and Monfardini]{scarselli2009graph}
Franco Scarselli, Marco Gori, Ah~Chung Tsoi, Markus Hagenbuchner, and Gabriele Monfardini.
\newblock The graph neural network model.
\newblock \emph{{IEEE} Trans. Neural Networks}, 20\penalty0 (1):\penalty0 61--80, 2009.
\newblock \doi{10.1109/TNN.2008.2005605}.
\newblock URL \url{https://doi.org/10.1109/TNN.2008.2005605}.

\bibitem[Song et~al.(2024)Song, Yuan, and Yang]{song2024fmint}
Zezheng Song, Jiaxin Yuan, and Haizhao Yang.
\newblock Fmint: Bridging human designed and data pretrained models for differential equation foundation model, 2024.
\newblock URL \url{https://arxiv.org/abs/2404.14688}.

\bibitem[Sperduti and Starita(1997)]{sperduti1997supervised}
Alessandro Sperduti and Antonina Starita.
\newblock Supervised neural networks for the classification of structures.
\newblock \emph{{IEEE} Trans. Neural Networks}, 8\penalty0 (3):\penalty0 714--735, 1997.
\newblock \doi{10.1109/72.572108}.
\newblock URL \url{https://doi.org/10.1109/72.572108}.

\bibitem[Takamoto et~al.(2022)Takamoto, Praditia, Leiteritz, MacKinlay, Alesiani, Pfl{\"{u}}ger, and Niepert]{takamoto2022pdebench}
Makoto Takamoto, Timothy Praditia, Raphael Leiteritz, Daniel MacKinlay, Francesco Alesiani, Dirk Pfl{\"{u}}ger, and Mathias Niepert.
\newblock Pdebench: An extensive benchmark for scientific machine learning.
\newblock In \emph{Advances in Neural Information Processing Systems 35: Annual Conference on Neural Information Processing Systems 2022, NeurIPS 2022, New Orleans, LA, USA, November 28 - December 9, 2022}, 2022.

\bibitem[Ulyanov et~al.(2016)Ulyanov, Vedaldi, and Lempitsky]{ulyanov2016instance}
Dmitry Ulyanov, Andrea Vedaldi, and Victor~S. Lempitsky.
\newblock Instance normalization: The missing ingredient for fast stylization.
\newblock \emph{CoRR}, abs/1607.08022, 2016.
\newblock URL \url{http://arxiv.org/abs/1607.08022}.

\bibitem[Wang et~al.(2024)Wang, Seidman, Sankaran, Wang, and Paris]{wang2024cvit}
Sifan Wang, Jacob~H Seidman, Shyam Sankaran, Hanwen Wang, and George J~Pappas Paris.
\newblock Cvit: Continuous vision transformer for op-erator learning.
\newblock \emph{arXiv preprint arXiv:2405.13998}, 3, 2024.

\bibitem[Weiler and Cesa(2019)]{weiler2019general}
Maurice Weiler and Gabriele Cesa.
\newblock General e(2)-equivariant steerable cnns.
\newblock In \emph{Advances in Neural Information Processing Systems 32 (NeurIPS 2019)}, 2019.

\bibitem[Xu et~al.(2023)Xu, Ye, Yan, Shi, Ye, Xu, Li, Bi, Qian, Wang, Xu, Zhang, Huang, Huang, and Zhou]{xu2023mplug2}
Haiyang Xu, Qinghao Ye, Ming Yan, Yaya Shi, Jiabo Ye, Yuanhong Xu, Chenliang Li, Bin Bi, Qi~Qian, Wei Wang, Guohai Xu, Ji~Zhang, Songfang Huang, Fei Huang, and Jingren Zhou.
\newblock mplug-2: {A} modularized multi-modal foundation model across text, image and video.
\newblock In \emph{International Conference on Machine Learning, {ICML} 2023, 23-29 July 2023, Honolulu, Hawaii, {USA}}, volume 202 of \emph{Proceedings of Machine Learning Research}, pages 38728--38748. {PMLR}, 2023.

\bibitem[Zaheer et~al.(2017)Zaheer, Kottur, Ravanbakhsh, P{\'{o}}czos, Salakhutdinov, and Smola]{zaheer2017deep}
Manzil Zaheer, Satwik Kottur, Siamak Ravanbakhsh, Barnab{\'{a}}s P{\'{o}}czos, Ruslan Salakhutdinov, and Alexander~J. Smola.
\newblock Deep sets.
\newblock In \emph{Advances in Neural Information Processing Systems 30: Annual Conference on Neural Information Processing Systems 2017, December 4-9, 2017, Long Beach, CA, {USA}}, 2017.

\bibitem[Zeiler et~al.(2010)Zeiler, Krishnan, Taylor, and Fergus]{zeiler2010deconvolutional}
Matthew~D. Zeiler, Dilip Krishnan, Graham~W. Taylor, and Rob Fergus.
\newblock Deconvolutional networks.
\newblock In \emph{2010 IEEE Computer Society Conference on Computer Vision and Pattern Recognition}, pages 2528--2535, 2010.
\newblock \doi{10.1109/CVPR.2010.5539957}.

\bibitem[Zhou et~al.(2024)Zhou, Ma, Wu, Wang, and Long]{zhou2024unisolver}
Hang Zhou, Yuezhou Ma, Haixu Wu, Haowen Wang, and Mingsheng Long.
\newblock Unisolver: Pde-conditional transformers are universal pde solvers, 2024.
\newblock URL \url{https://arxiv.org/abs/2405.17527}.

\bibitem[Zhou et~al.(2020)Zhou, Cui, Hu, Zhang, Yang, Liu, Wang, Li, and Sun]{zhou2020graph}
Jie Zhou, Ganqu Cui, Shengding Hu, Zhengyan Zhang, Cheng Yang, Zhiyuan Liu, Lifeng Wang, Changcheng Li, and Maosong Sun.
\newblock Graph neural networks: {A} review of methods and applications.
\newblock \emph{{AI} Open}, 1:\penalty0 57--81, 2020.
\newblock \doi{10.1016/J.AIOPEN.2021.01.001}.
\newblock URL \url{https://doi.org/10.1016/j.aiopen.2021.01.001}.

\end{thebibliography}

\newpage
\section*{NeurIPS Paper Checklist}

\begin{enumerate}

\item {\bf Claims}
    \item[] Question: Do the main claims made in the abstract and introduction accurately reflect the paper's contributions and scope?
    \item[] Answer: \answerYes{} 
    \item[] Justification: The abstract and introduction reflects our main contribution: novel architecture for dimension-free training and inference.

\item {\bf Limitations}
    \item[] Question: Does the paper discuss the limitations of the work performed by the authors?
    \item[] Answer: \answerYes{} 
    \item[] Justification: The paper discuss the limitations in the last paragraph of the conclusion.

\item {\bf Theory assumptions and proofs}
    \item[] Question: For each theoretical result, does the paper provide the full set of assumptions and a complete (and correct) proof?
    \item[] Answer: \answerYes{} 
    \item[] Justification: The paper provides the assumptions and proofs in the appendix.

    \item {\bf Experimental result reproducibility}
    \item[] Question: Does the paper fully disclose all the information needed to reproduce the main experimental results of the paper to the extent that it affects the main claims and/or conclusions of the paper (regardless of whether the code and data are provided or not)?
    \item[] Answer: \answerYes{} 
    \item[] Justification: The paper provides experimental details in the experiment section and the appendix.

\item {\bf Open access to data and code}
    \item[] Question: Does the paper provide open access to the data and code, with sufficient instructions to faithfully reproduce the main experimental results, as described in supplemental material?
    \item[] Answer: \answerYes{} 
    \item[] Justification: We share the source code with configuration files in the submission and the data is publicly available.

\item {\bf Experimental setting/details}
    \item[] Question: Does the paper specify all the training and test details (e.g., data splits, hyperparameters, how they were chosen, type of optimizer, etc.) necessary to understand the results?
    \item[] Answer: \answerYes{} 
    \item[] Justification: The paper provides the experiemental details including hyperparameters in the appendix section.

\item {\bf Experiment statistical significance}
    \item[] Question: Does the paper report error bars suitably and correctly defined or other appropriate information about the statistical significance of the experiments?
    \item[] Answer: \answerNo{} 
    \item[] Justification: Error bars are not reported because the pretraining is too computationally expensive.

\item {\bf Experiments compute resources}
    \item[] Question: For each experiment, does the paper provide sufficient information on the computer resources (type of compute workers, memory, time of execution) needed to reproduce the experiments?
    \item[] Answer: \answerYes{} 
    \item[] Justification: The paper provide such information in the appendix.
    
\item {\bf Code of ethics}
    \item[] Question: Does the research conducted in the paper conform, in every respect, with the NeurIPS Code of Ethics \url{https://neurips.cc/public/EthicsGuidelines}?
    \item[] Answer: \answerYes{} 
    \item[] Justification: The paper only targets mathematical data, which does not involve any harmful aspect.

\item {\bf Broader impacts}
    \item[] Question: Does the paper discuss both potential positive societal impacts and negative societal impacts of the work performed?
    \item[] Answer: \answerNA{} 
    \item[] Justification: The paper only targets mathematical data, which does not involve any societal impacts.
    
\item {\bf Safeguards}
    \item[] Question: Does the paper describe safeguards that have been put in place for responsible release of data or models that have a high risk for misuse (e.g., pretrained language models, image generators, or scraped datasets)?
    \item[] Answer: \answerNA{} 
    \item[] Justification: The paper trains with only PDEs.

\item {\bf Licenses for existing assets}
    \item[] Question: Are the creators or original owners of assets (e.g., code, data, models), used in the paper, properly credited and are the license and terms of use explicitly mentioned and properly respected?
    \item[] Answer: \answerYes{} 
    \item[] Justification: The paper clarifies the source of baseline codes and data.

\item {\bf New assets}
    \item[] Question: Are new assets introduced in the paper well documented and is the documentation provided alongside the assets?
    \item[] Answer: \answerYes{} 
    \item[] Justification: The paper will release the code with the documentation after accept.

\item {\bf Crowdsourcing and research with human subjects}
    \item[] Question: For crowdsourcing experiments and research with human subjects, does the paper include the full text of instructions given to participants and screenshots, if applicable, as well as details about compensation (if any)? 
    \item[] Answer: \answerNA{} 
    \item[] Justification: No crowdsourcing and human subjects.

\item {\bf Institutional review board (IRB) approvals or equivalent for research with human subjects}
    \item[] Question: Does the paper describe potential risks incurred by study participants, whether such risks were disclosed to the subjects, and whether Institutional Review Board (IRB) approvals (or an equivalent approval/review based on the requirements of your country or institution) were obtained?
    \item[] Answer: \answerNA{} 
    \item[] Justification: No crowdsourcing and human subjects.

\item {\bf Declaration of LLM usage}
    \item[] Question: Does the paper describe the usage of LLMs if it is an important, original, or non-standard component of the core methods in this research? Note that if the LLM is used only for writing, editing, or formatting purposes and does not impact the core methodology, scientific rigorousness, or originality of the research, declaration is not required.
    \item[] Answer: \answerNA{} 
    \item[] Justification: LLM is used only for editing.

\end{enumerate}

\newpage
\appendix
\section{Dimension-Agnostic PDE Solver}
\label{app:sec:pde_solver}

\begin{figure}[t]
    \centering
    \begin{subfigure}[b]{0.33\textwidth}
    \centering
    \includegraphics[width=\linewidth]{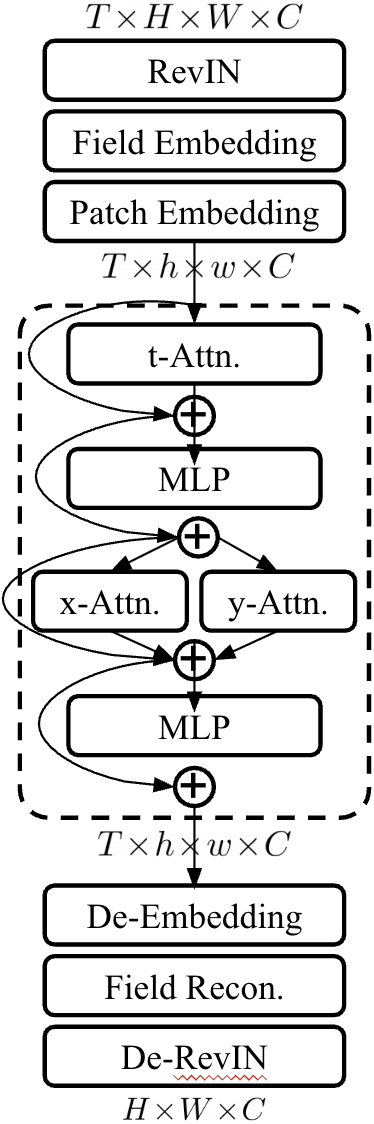}
    \caption{MPP}
    \label{fig:mpp}
    \end{subfigure}%
    \hfill
    \begin{subfigure}[b]{0.61\textwidth}
    \centering
    \includegraphics[width=\linewidth]{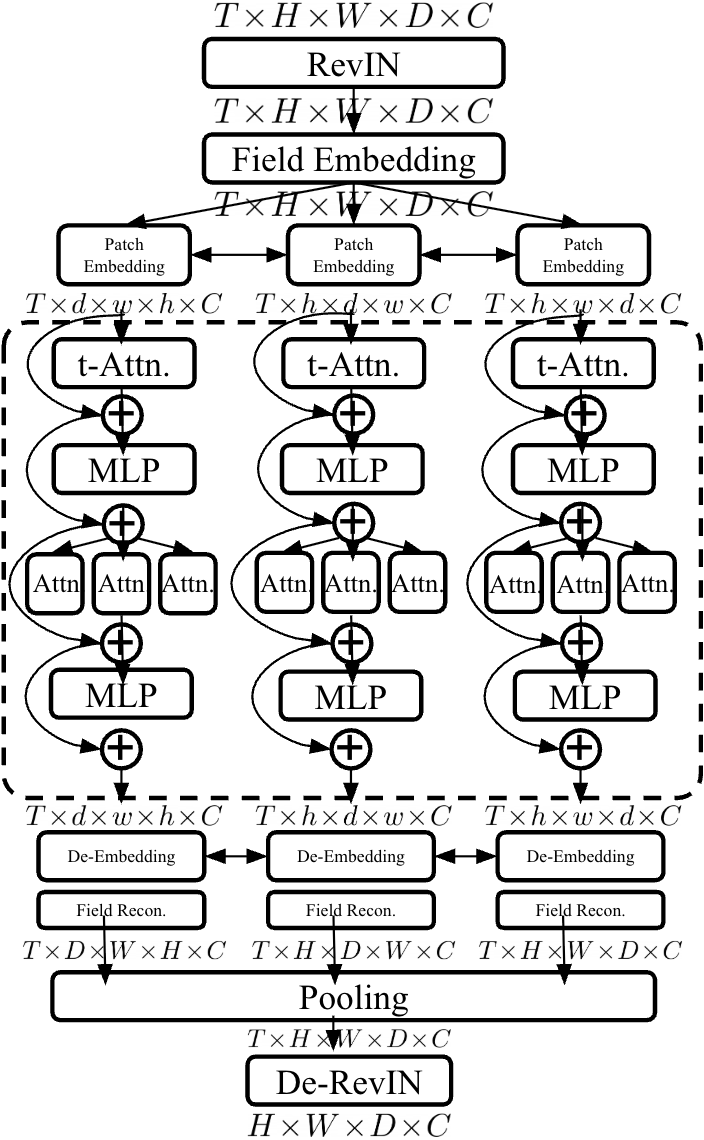}
    \caption{Dimension-Agnostic MPP}
    \label{fig:xmpp}
    \end{subfigure}
    \caption{Comparison between MPP and its dimension-agnostic version.}
    \label{fig:mpp_vs_xmpp}
\end{figure}

In this section, we present an example of the dimension-agnostic PDE solver using the ViT-based model MPP as a backbone. MPP processes a PDE solution from timestep 1 to $T$ (e.g., $\bsx_{1:T} \in \bbR^{T \times H \times W \times C}$) and outputs the next timestep $T+1$ (e.g., $\bsx_{T+1} \in \bbR^{H \times W \times C}$). MPP consists of patch embedding, multiple self-attention layers, and patch de-embedding. In the self-attention layers, MPP separates the attention along the time axis (temporal self-attention) in a depthwise manner. We also follow this separate temporal self-attention, but make the spatial self-attention dimension-agnostic. The comparison between MPP and its dimension-agnostic version is illustrated in~\cref{fig:mpp_vs_xmpp}.

\textbf{Patch Embedding.}
The patch embedding is a three-layer CNN, used as the lifting layer. The original patch embedding in MPP is as follows:
\[
\begin{aligned}
&\operatorname{EmbBlock}(\bsx) = \operatorname{GeLU}(\operatorname{Norm}(\operatorname{Conv2D}(\bsx))),\\
&\operatorname{Embedder}(\bsx) = \operatorname{Conv2D}(\operatorname{EmbBlock}(\operatorname{EmbBlock}(\bsx))),\\
&\bsh = \operatorname{Norm}(\operatorname{Embedder}(\bsx)),
\end{aligned}
\]
where $\operatorname{Norm}$ indicates InstanceNorm~\citep{ulyanov2016instance}. On the other hand, in the axial version, we use $\operatorname{Embedder}$ for $\psi$ in~\cref{eq:lifting-k} with $\max$ aggregation as follows:
\[
\begin{aligned}
&\tilde{\bsh}_{d_i} = \max_{j \neq i,\ j \in [1,K]} \Tr{(K-1)}{(j)} \left( \operatorname{AvgPool}_{1,\ldots,K-2} \left( \operatorname{Embedder}_{K-1,K} \left( \Tr{}{(j)(K-1)(i)(K)}(\bsx) \right) \right) \right),\\
&\bsh_{d_i} = \operatorname{Norm}(\tilde{\bsh}_{d_i}),
\end{aligned}
\]
where $\Tr{(m)}{(n)}(\bsx)$ denotes the transpose between the $m$-th and $n$-th axes. Through $\Tr{(j)}{(K-1)}\Tr{(i)}{(K)}$, the $j$-th and $i$-th axes are located at the $(K-1)$-th and $K$-th positions, respectively. Then, $\operatorname{Embedder}_{K-1,K}$ applies a 2D CNN along the $K-1$ and $K$ axes, and the remaining axes $1,\ldots,K-2$ are reduced via average pooling to match the output tensor size. Since the final $\operatorname{Norm}$ is a pointwise operation, it preserves axis-permutation equivariance. We replace InstanceNorm with LayerNorm~\citep{ba2016layer} to avoid training instability in multidimensional training.

\textbf{Self-Attentions.}
MPP already employs set-based axial self-attention, which outputs a single feature as follows:
\[
\tilde{\bsh}' = \frac{1}{K} \sum_{i=1}^K \Tr{(K)}{(i)}(\operatorname{SelfAttn}_K(\Tr{(i)}{(K)}(\bsh))), \quad
\bsh' = \bsh + \tilde{\bsh}'.
\]
In contrast, in the graph-based axial case, we obtain $K$ features due to the lifting layer in the patch embedding. Therefore, we also use a set-based structure but apply the same attention to each feature independently, which differs from the graph-based attention example in~\cref{eq:subsequent-attn2}. The self-attention used is:
\[
\tilde{\bsh}'_{d_i} = \frac{1}{K} \sum_{i=1}^K \Tr{(K)}{(i)}(\operatorname{SelfAttn}_K(\Tr{(i)}{(K)}(\bsh_{d_i}))), \quad
\bsh'_{d_i} = \bsh_{d_i} + \tilde{\bsh}'_{d_i}, \quad \forall i \in [1,K].
\] Additional operations such as layer normalization and drop residual paths are included, but omitted here for simplicity. These operations follow MPP exactly.

\textbf{Patch De-Embedding.}
In this block, we recover the input tensor's width and height using a CNN consisting of $\operatorname{ConvTranspose}$ layers (also known as deconvolutions)~\citep{zeiler2010deconvolutional}. In the original implementation, this CNN is defined as:
\[
\begin{aligned}
\operatorname{DeembBlock}(\bsh') &= \operatorname{GeLU}(\operatorname{Norm}(\operatorname{ConvT2D}(\bsh'))),\\
\operatorname{Deembedder}(\bsh') &= \operatorname{ConvT2D}(\operatorname{DeembBlock}(\operatorname{DeembBlock}(\bsh'))),\\
\bsh'' &= \operatorname{Deembedder}(\bsh').
\end{aligned}
\]
In the axial version, we define two $\operatorname{Deembedder}$ blocks: one for updating the node feature and the other for updating the neighborhood feature (nhbr). We aggregate them using a $\max$ operation. Then, \cref{eq:subsequent-k} becomes:
\[
\begin{aligned}
{\bsh''}_{d_i}^{\text{(node)}} &= \operatorname{Deembedder}_{K-1,K}^{\text{(node)}}(\bsh'_{d_i}),\\
{\bsh''}_{d_i}^{\text{(nhbr)}} &= \max_{j \neq i,\, j \in [1,K]} \Tr{(i)}{(K)} \Tr{(K)}{(j)} \operatorname{Deembedder}_{K-1,K}^{\text{(nhbr)}}(\bsh'_{d_j}),\\
{\bsh''}_{d_i} &= \max\left\{{\bsh''}_{d_i}^{\text{(node)}}, {\bsh''}_{d_i}^{\text{(nhbr)}}\right\}, \quad \forall i \in [1,K],
\end{aligned}
\]
where $\Tr{(i)}{(K)} \Tr{(K)}{(j)}$ aligns the axis order of $\bsh'_{d_j}$ to match that of $\bsh'_{d_i}$.

Unfortunately, this dimension-agnostic PDE solver is not axis-permutation equivariant because the \emph{positional encoding} (not mentioned in this section but present in the actual implementation) and the patch de-embedding layer break axis-permutation equivariance, even though the solver anyway works for every dimensionality.
\section{Proofs of Theorems}
\label{app:sec:proofs}

\ActivateWarningFilters[pdftoc]
\subsection{\cref{thm:sxnn}}
\DeactivateWarningFilters[pdftoc]
\label{app:sec:sxnn-proof}

\emph{
    Let $\phi$ be an axis-permutation equivariant function (e.g., another SXNN or pointwise operation). The SXNN in~\cref{eq:sxnn} is axis-permutation equivariant for any rank-$K$ tensor $\bsx \in \bbR^{d_1 \Times d_2 \Times \ldots \Times d_K}$.
}
\begin{proof}
We prove the axis-permutation equivariance property defined in~\cref{eq:axis-perm-equiv}. For any axis permutation $\Pi' \in \bPi$, the permutation of the input in~\cref{eq:sxnn} before applying $\phi$ becomes
\[
\bigoplus_{i=1}^{K!-1} \Pi_i^{-1}\bigl(\psi(\Pi_i(\Pi'(\bsx)))\bigr).
\]
Now, by substituting $\Pi_j = \Pi_i \Pi'$ (equivalently, $\Pi_i = \Pi_j {\Pi'}^{-1}$), we get
\[
\begin{aligned}
\bigoplus_{i=1}^{K!-1} \Pi_i^{-1}\bigl(\psi(\Pi_i({\Pi'}(\bsx)))\bigr)
&= \bigoplus_{j=1}^{K!-1} {\Pi'} \Pi_j^{-1} \bigl(\psi(\Pi_j(\bsx))\bigr) \\
&= {\Pi'} \bigoplus_{j=1}^{K!-1} \Pi_j^{-1} \bigl(\psi(\Pi_j(\bsx))\bigr) \\
&= {\Pi'} \bigoplus_{i=1}^{K!-1} \Pi_i^{-1} \bigl(\psi(\Pi_i(\bsx))\bigr).
\end{aligned}
\]
The last equality holds due to the permutation-invariant nature of the operation $\bigoplus$.

By the assumption that $\phi$ is axis-permutation equivariant, it follows that
\[
\begin{aligned}
\phi\bigg(\bigoplus_{i=1}^{K!-1} \Pi_i^{-1}\bigl(\psi(\Pi_i({\Pi'}(\bsx)))\bigr)\bigg)
&= \phi\bigg({\Pi'} \bigoplus_{i=1}^{K!-1} \Pi_i^{-1} \bigl(\psi(\Pi_i(\bsx))\bigr)\bigg) \\
&= {\Pi'} \phi\bigg(\bigoplus_{i=1}^{K!-1} \Pi_i^{-1} \bigl(\psi(\Pi_i(\bsx))\bigr)\bigg),
\end{aligned}
\]
which proves that~\cref{eq:sxnn} is axis-permutation equivariant for all ${\Pi'} \in \bPi$.
\end{proof}

\ActivateWarningFilters[pdftoc]
\subsection{\cref{thm:lifting}}
\DeactivateWarningFilters[pdftoc]
\label{app:sec:lifting-proof}

\begin{assm}
\vspace{5pt}
\label{thm:assumption1}
All layers follow the same axis order. For example, if the lifting generates $\bsh_H$ and $\bsh_W$ with shapes $H\Times W\Times C$ and $W\Times H\Times C$, respectively, then every layer should produce $\bsh'_H$ and $\bsh'_W$ with the same shapes.
\vspace{5pt}
\end{assm}

\begin{assm}
\vspace{5pt}
\label{thm:assumption2}
In the lifting layer, $\phi$ must be axis-permutation equivariant along axes $1,2,\ldots,K-1$, and $\psi$ along axes $1,2,\ldots,K-2$. For instance, if $\bsx\in\bbR^{H\Times W\Times D\Times C}$ for $K=3$, then $\Tr{1}{2}(\phi(\bsx)) = \phi(\Tr{1}{2}(\bsx))$ must hold, but $\Tr{1}{3}(\phi(\bsx)) = \phi(\Tr{1}{3}(\bsx))$ is not required.
\vspace{5pt}
\end{assm}

Under assumptions \cref{thm:assumption1} and \cref{thm:assumption2},
the lifting layer of GXNN, \cref{eq:lifting-k}, is axis-permutation equivariant for any rank-$K$ (except $K=1$) tensor $\bsx\in\bbR^{d_1\Times d_2\Times\ldots\Times d_K}$.

\begin{proof}
The lifting layer $\operatorname{Lifting}$ described in \cref{eq:lifting-k} can be equivalently written as
\[
\bsh_{d_i} = \phi\bigg(\Tr{(i)(K)}{}(\bsx), \bigoplus_{j\neq i}^K \Tr{(i)(K)}{}\Tr{(i)(K)(j)(K-1)}{}^{-1}\psi\left(\Tr{(i)(K)(j)(K-1)}{}(\bsx)\right)\bigg)
\] where $\Tr{(a)(b)(c)(d)}{}=\Tr{abcd}{}$ denotes the axes permutation from indices $\{1,2,3,4\}$ to $\{a,b,c,d\}$ as explained in~\cref{sec:main:cycle}.
Applying an axis permutation $\Pi(\bsx)$ to the input gives
\[
\phi\bigg(\Tr{(i)(K)}{}(\Pi(\bsx)), \bigoplus_{j\neq i}^K \Tr{(i)(K)}{}\Tr{(i)(K)(j)(K-1)}{}^{-1}\psi\left(\Tr{(i)(K)(j)(K-1)}{}(\Pi(\bsx))\right)\bigg)
\]
Substituting $\Pi'\Tr{(k)}{(K)} = \Tr{(i)}{(K)}\Pi$ for $i = \pi(k)$ and $\Pi''\Tr{(i)(K)(j)(K-1)}{} = \Tr{(k)(K)(l)(K-1)}{}\Pi$ for $i = \pi(k),j=\pi'(l)$, where $\Pi'$ permutes only axes $1$ through $K-1$ and $\Pi''$ permutes only axes $1$ through $K-2$, we get
\[
= \phi\bigg(\Pi'\Tr{(k)(K)}{}(\bsx), \bigoplus_{\pi'(l)\neq\pi(k)}^K \Pi'\Tr{(k)(K)}{}\Tr{(k)(K)(l)(K-1)}{}^{-1}{\Pi''}^{-1}\psi\left(\Pi''\Tr{(k)(K)(l)(K-1)}{}(\bsx)\right)\bigg).
\]
Let $\pi,\pi'$ be the element-wise permutation across axes induced by $\Pi$. Since the aggregation is permutation-invariant and $\pi'(l)\neq\pi(k)$ is equivalent to $l\neq k$ (as $\pi'$ does not act on the $K$-th axis and $k$-th axis is transposed to $K$ by $\Tr{(k)}{(K)}$), and using \cref{thm:assumption2} (permutation equivariance of $\psi$ under $\Pi''$), we have
\[
= \phi\bigg(\Pi'\Tr{(k)(K)}{}(\bsx), \bigoplus_{l\neq k}^K \Pi'\Tr{(k)(K)}{}\Tr{(k)(K)(l)(K-1)}{}^{-1}{\Pi''}^{-1}\Pi''\psi\left(\Tr{(k)(K)(l)(K-1)}{}(\bsx)\right)\bigg).
\]
Applying \cref{thm:assumption2} again for $\phi$ under $\Pi'$ yields
\[
\label{eq:lifting-equiv}
= \Pi'\phi\bigg(\Tr{(k)(K)}{}(\bsx), \bigoplus_{l\neq k}^K \Tr{(k)(K)}{}\Tr{(k)(K)(l)(K-1)}{}^{-1}\psi\left(\Tr{(k)(K)(l)(K-1)}{}(\bsx)\right)\bigg),
\] which follows the definition of axes-permutation equivariance as in~\cref{eq:axis-perm-equiv}.
\end{proof}

\ActivateWarningFilters[pdftoc]
\subsection{\cref{thm:subsequent}}
\DeactivateWarningFilters[pdftoc]
\label{app:sec:subsequent-proof}

\begin{assm}
\vspace{5pt}
\label{thm:assumption3}
In the subsequent layers, both $\phi$ and $\psi$ must be axis-permutation equivariant along axes $1,2,\ldots,K-1$.
\vspace{5pt}
\end{assm}

 Under assumptions \cref{thm:assumption1} and \cref{thm:assumption3}, the subsequent layers of GXNN in \cref{eq:subsequent-k} are axis-permutation equivariant for any rank-$K$ tensor $\bsh_{d_i}\in\bbR^{d_1\Times d_2\Times\ldots\Times d_K}$.

\begin{proof}
The subsequent layers $\operatorname{Subsequent}$ described in \cref{eq:subsequent-k} can be equivalently written as
\[
\bsh'_{d_i} = \phi\left(\bsh_{d_i}, \Tr{(i)}{(K)}\sum_{j=1}^K\Tr{(j)}{(K)}^{-1}\psi(\bsh_{d_j})\right).
\]
Using \cref{eq:lifting-equiv}, permutation of input leads to permutation of output of the lifting layer, which is the permutation of input of subsequent layers,
\[
\phi\left(\Pi'\bsh_{d_k}, \Tr{(\pi'(k))}{(K)}\sum_{\pi'(l)=1}^K\Tr{(\pi'(l))}{(K)}^{-1}\psi(\Pi'\bsh_{d_l})\right).
\]
Because the index exchanges in $\Tr{}{}$ must adjust to the input permutation $\Pi'$, and both $\phi$ and $\psi$ are $\Pi'$-equivariant by \cref{thm:assumption3}, and the sum is permutation-invariant:
\[
\label{eq:subsequent-equiv}
= \Pi'\phi\left(\bsh_{d_k}, \Tr{(k)}{(K)}\sum_{l=1}^K\Tr{(l)}{(K)}^{-1}\psi(\bsh_{d_l})\right)
= \Pi'\bsh'_{d_k}.
\]

\subsection{Axes-Permutation Equivariance of Pooling Layer}
Finally, the pooling layer aggregates the axis-wise features into a single feature, which remains equivariant to axis permutations $\Pi$. As described in the example in~\cref{eq:pooling}, it is equivalently written as
\[
\bsh'' = \sum_{j=1}^K\Tr{(j)}{(K)}^{-1}\bsh'_{d_j}.
\]
By definition,
\[
\Pi'\Tr{(k)}{(K)} = \Tr{(\pi(k))}{(K)}\Pi
\quad \Leftrightarrow \quad \Pi'\Tr{(K)}{(k)} = \Tr{(K)}{(\pi(k))}\Pi.
\]
Using this identity and \cref{eq:subsequent-equiv}, permutation of the input yields
\[
\sum_{\pi'(l)=1}^K\Tr{(\pi'(l))}{(K)}^{-1}\Pi'\bsh'_{d_l}
&= \sum_{\pi'(l)=1}^K\Pi'\Tr{(l)}{(K)}^{-1}\bsh'_{d_l}
= \sum_{\pi'(l)=1}^K\Tr{(\pi(l))}{(K)}^{-1}\Pi\bsh'_{d_l}\\
&= \sum_{\pi'(l)=1}^K\Pi\Tr{(l)}{(K)}^{-1}\bsh'_{d_l}
= \Pi\sum_{l=1}^K\Tr{(l)}{(K)}^{-1}\bsh'_{d_l}
= \Pi\bsh'',
\] which concludes that the axis-permutation of the input yields the axis-permutation of the output.
\end{proof}

\section{X-CViT and X-MPP}
\label{app:sec:variant}

For X-MPP, we exactly followed the example of the dimension-agnostic PDE solver introduced in~\cref{app:sec:pde_solver}.

For X-CViT, we mostly followed the structure of CViT but modified the dimension-dependent components, including patch embedding, spatial self-attention, coordinate query embedding, and decoder cross-attention. For the remaining components, we refer the reader to~\citep{wang2024cvit}.

\textbf{Patch Embedding.}
X-CViT also uses patch embedding as a lifting layer. In the original CViT, the patch embedding is defined as
\[
\bsh = \operatorname{Conv2D}(\bsx).
\]
In contrast, in X-CViT, the patch embedding is defined as
\[
\bsh_{d_i} = \frac{1}{K}\sum_{j\neq i}^K \left(\Tr{(K-1)}{(j)(K)(i)}\left(\operatorname{MaxPool}_{1,\ldots,K-2}\left(\operatorname{Conv2D}_{K-1,K}\left(\Tr{(i)(K)(j)}{(K-1)}(\bsx)\right)\right)\right)\right),
\]
where $\Tr{(i)(K)(j)}{(K-1)}$ is used to apply $\operatorname{Conv2D}_{K-1,K}$ along the $i$-th and $j$-th axes, and $\Tr{(K-1)(j)(K)(i)}{}$ reorders the axes back to enable aligned feature aggregation through the summation.

\textbf{Spatial Self-Attentions.}
CViT separates temporal and spatial attention and X-CViT also follows the separated structure. In X-CViT, we use set-based axial self-attention for each feature. The original self-attention in CViT is defined as
\[
\bsh' = \operatorname{SelfAttn}(\operatorname{Flatten}(\bsx)),
\]
which flattens the spatial axes of the input into a single sequence before applying self-attention (e.g., $H\Times W\to HW$). In contrast, X-CViT does not perform flattening but instead aggregates self-attention outputs along each axis:
\[
\bsh'_{d_i} = \bsh_{d_i} + \sum_{j=1}^K \Tr{(K)}{(j)}\left(\operatorname{SelfAttn}(\Tr{(j)}{(K)}(\bsh_{d_i}))\right),
\]
where $\Tr{(j)}{(K)}$ is used to apply $\operatorname{SelfAttn}$ along the $j$-th axis, and $\Tr{(K)}{(j)}$ reorders the axes back for aggregation.

\textbf{Pooling Axial Features.}
Right after computing the encoder, we still have $K$ axial features lifted by the patch embedding. Thus, before applying the decoder, we aggregate them by averaging to obtain a single feature.

\textbf{Coordinate Query Embedding.}
While MPP always predicts the entire spatial grid points at the next timestep, CViT selects where to predict by providing spatial coordinates, assuming the PDE solutions lie on a $[0,1]^2$ grid. To achieve this, CViT calculates the distance between $N$ query coordinates and all grid points in $G^K$, resulting in a tensor of shape $N \times G^K$, which is then encoded into a $N \times Q$ tensor, where $Q$ is the feature size of the coordinate embedding. CViT uses two linear layers to embed $G^K$ into $\bbR^Q$ for the distance tensor $\bsq \in \bbR^{N \times G^K}$ (with $K=2$ in 2D), treating $N$ as a batch dimension:
\[
\bsq' = \operatorname{Norm}(\operatorname{Linear}(\operatorname{Linear}(\bsq))).
\]

On the other hand, X-CViT utilizes a set-based axial linear layer with $\max$ aggregation as follows:
\[
\bsq' = \max_{i\in[1:K]}\Tr{(K)}{(i)}\left(\operatorname{Norm}_K\left(\operatorname{Linear}_K\left(\operatorname{GlobalMaxPool}_{1,\ldots,K-1}\left(\operatorname{Linear}_K(\Tr{(i)}{(K)}(\operatorname{Unflatten}(\bsq)))\right)\right)\right)\right),
\]
where $\operatorname{Unflatten}$ reshapes $G^K \to G \times G \times \cdots \times G$, and $\Tr{(i)}{(K)}$ is used to apply $\operatorname{Linear}_K$ along the $i$-th axis. Note that $\operatorname{GlobalMaxPool}_{1,\ldots,K-1}$ reduces the size of axes $1$ through $K-1$ to one, converting $\bbR^{Q \times \cdots \times Q}$ to $\bbR^Q$. This illustrates that XNN can be applied even when the number of input axes differs from the number of output axes.

\textbf{Decoder Cross-Attention.}
In the decoder cross-attention, we perform attention by treating the coordinate embedding $\bsq$ as the query and the features produced by the CViT encoder as the key and value. Since this operation involves a rank-1 vector and a rank-$K$ tensor, we must carefully design the XNN architecture to handle such multi-dimensional cross-attention. In CViT, the cross-attention is formulated as
\[
\bsh'' = \bsq' + \operatorname{Attn}(\bsq', \operatorname{Flatten}(\bsh'), \operatorname{Flatten}(\bsh')).
\]
In contrast, X-CViT uses SXNN with a $\operatorname{Repeat}$ function to align dimensions and tensor sizes:
\[
\bsh'' = \bsq' + \sum_{i=1}^K\operatorname{GlobalMaxPool}_{1,\ldots,K-1}\left(\operatorname{Attn}(\operatorname{Repeat}_{1,\ldots,K-1}(\bsq'), \Tr{(i)}{(K)}(\bsh'), \Tr{(i)}{(K)}(\bsh'))\right),
\]
where $\operatorname{Repeat}_{1,\ldots,K-1}(\bsq')$ expands and repeats the vector $\bsq' \in \bbR^{Q}$ along axes $1,\ldots,K-1$, resulting in $\operatorname{Repeat}_{1,\ldots,K-1}(\bsq') \in \bbR^{Q \times Q \times \cdots \times Q}$. 
\section{Experimental Details}
\label{app:sec:detail}

\subsection{GP Kernel Prediction}
\textbf{Synthetic Dataset Construction.} 
To construct the synthetic dataset, samples were generated from zero-mean Gaussian processes using either the radial basis function kernel or the periodic kernel.
For each kernel, data were generated in dimensions ranging from 2 to 5.
In each dimension $d \in \{2, 3, 4, 5\}$, a structured grid that equally divides the hypercube $[-2, 2]^d$ was created over the input space, with grid sizes tailored to $2^{7-d}$.
This results in output shapes as specified in~\cref{tab:tensor_shapes}, where the last channel of length 1 was appended for convenience.

\begin{table}[h]
    \caption{Shape of tensors in the synthetic dataset}
    \label{tab:tensor_shapes}
    \centering
    \begin{tabular}{cccc}
        \toprule
         Dimensions & Shape & Dimensions & Shape \\
         \midrule
         2D & (32, 32, 1) & 4D & (8, 8, 8, 8, 1) \\
         \midrule
         3D & (16, 16, 16, 1) & 5D & (4, 4, 4, 4, 4, 1) \\
         \bottomrule
    \end{tabular}
\end{table}

Each grid point represents an input to the GP kernel, from which the full covariance matrix was computed for a batch of samples. 
Kernel parameters were randomly sampled, i.e., length within $[0.1, 0.6]$, scale within $[0.1, 1]$, and period within $[0.1, 0.5]$ for the periodic kernel, to introduce variation.
The resulting covariance matrices were corrected via eigenvalue clipping to ensure positive semidefiniteness.
Each generated sample from the periodic kernel was labeled 0, while those from the RBF kernel were labeled 1. 
An equal number of samples from each kernel type was used to ensure balanced classes.
All data were normalized to have zero mean and unit variance before being fed into the models. 
The dataset was then divided into training and validation subsets with an 80/20 split.

\textbf{3D CNN.} 
As 3D convolution operations are only compatible with 5D inputs, we forced the dimensions of the tensor to 5; for lower-dimensional samples, we zero-padded along missing axes; for higher-dimensional samples, we reshaped the tensor to be 5D by aggregating all the dimensions after the third dimension.
We adopted a conventional 3D convolutional neural network operating on tensors of shape $(B,H,W,D,C)$, where $B$ is the batch size.
The network consists of three stacked 3D convolutional layers, each with 32 filters and kernel sizes of $3 \times 3 \times3$, followed by LayerNorm and ReLU activations. 
After the convolutional layers, global max pooling is applied across all spatial dimensions, and a final dense layer maps the resulting features to a single output for binary classification.

\textbf{SXCNN.} Set-based XCNN generalizes convolution to multidimensional inputs using directional convolutions along each axis. At its core is the SXConv module, which applies 1D convolutions separately along each axis and merges the resulting features using an element-wise maximum. Each SXConv operation is followed by LayerNorm and a ReLU activation. The network consists of five SXConv layers, with each layer operating at 64 hidden features, effectively doubling the base hidden dimensionality. The final feature maps are globally pooled across all spatial dimensions, and classification is performed using a fully connected output layer.

\textbf{GXCNN.} Graph-based XCNN introduces a more complex interaction between spatial axes through the use of cross-axial convolutions. The architecture begins with a lifting layer, which performs 2D convolutions across each pair of axes to produce a set of intermediate representations. This is followed by a series of XConv layers, where each layer simultaneously considers pairs of axes using separate \textit{node} and \textit{neighbor} convolutions. The feature maps from each interaction are merged using max operations after appropriate axis permutation. Each convolution operation is followed by LayerNorm and ReLU activation. The network consists of one XLift layer and four XConv layers, each with 32 hidden features. Axial max pooling was applied across each dimension, followed by global max pooling, and the pooled output was passed to a dense layer for binary classification.

\textbf{Hyperparameters.} All models were trained using the Adam optimizer, with a fixed learning rate of 0.001 and a batch size of 64. Training was conducted for 10 epochs for each model. The loss function used was binary cross-entropy computed from the sigmoid of the output logits. Training and evaluation were implemented in JAX 0.4.30~\cite{bradbury2018jax} and Flax 0.8.5~\cite{heek2024flax}, with PyTorch 2.7.0+cu118~\cite{paszke2019pytorch} used primarily for data preprocessing and batching. To ensure that performance comparisons were valid, the same hyperparameters and preprocessing procedures were applied across all models.
The table below summarizes the hyperparameter settings used for training each of the three neural network models.
\begin{table}[h]
\centering
\caption{Hyperparameters for each model architecture}
\label{tab:training-hyperparams}
\begin{tabular}{llccccc}
\toprule
\textbf{Model} & \textbf{Architecture} & \textbf{Hidden Dim} & \textbf{Learning Rate} & \textbf{Batch Size} & \textbf{Epochs} \\
\midrule
CNN    & 3 Conv layers            & 32  & 1e-3 & 64 & 10 \\
SXCNN & 5 SXConv layers          & 64  & 1e-3 & 64 & 10 \\
XCNN   & 1 XLift + 4 XConv layers & 32  & 1e-3 & 64 & 10 \\
\bottomrule
\end{tabular}
\end{table}

\subsection{PDE Foundation Model}

\textbf{Hardware and Software.}
We implemented CViT and X-CViT using JAX 0.4.30~\citep{bradbury2018jax} and FLAX 0.8.5~\citep{heek2024flax}, while MPP and X-MPP were implemented using PyTorch 2.1.0+cu121~\citep{paszke2019pytorch}. All experiments were conducted on NVIDIA GPUs: RTX 3090, RTX A6000, and RTX 5090. For the RTX 5090 machine, we used PyTorch 2.8.0+cu128 due to CUDA driver compatibility.

\textbf{Training Loss.}
In CViT training, the objective function is the $l_2$ loss between the predicted next-timestep solution and the ground truth:
\[
\calL_\text{CViT} = \frac{1}{|B|}\sum_{\bsx\in\calX}\norm{\hat{\bsx}_{t+1} - f(\bsx_{(t-s):t})}_2^2,
\]
where $B$ is the mini-batch, $\hat{\bsx}_{t+1}$ is the ground-truth solution at the next timestep, $f$ is the neural PDE solver being trained, $t$ is the timestep, and $s$ is the number of input timesteps.

However, different PDEs exhibit varying magnitudes in their state variables, which can lead to imbalanced training that overemphasizes PDEs with larger magnitudes. To address this, \citet{mccabe2024multiple} used the normalized mean squared error (NMSE), which scales each output to unit magnitude. We adopted the same loss function for X-MPP training:
\[
\calL_\text{MPP} = \frac{1}{|B|}\sum_{\bsx\in\calX}\frac{\norm{\hat{\bsx}_{t+1} - f(\bsx_{(t-s):t})}_2^2}{\norm{\hat{\bsx}_{t+1}}_2^2 + \epsilon}.
\]

Since each PDE may produce tensors of different sizes, composing multiple PDEs in a single mini-batch is not feasible. Instead, we accumulate gradients across multiple mini-batches by summing them before performing a parameter update.

\textbf{Evaluation Metric.}
For evaluation, we use the normalized root mean squared error (NRMSE), which is the square root of the normalized mean squared error (NMSE), to compare with baseline methods. The metric is defined as
\[
\label{eq:nrmse}
\frac{1}{|B|}\sum_{\bsx\in\calX} \frac{\norm{\hat{\bsx}_{t+1} - f(\bsx_{(t-s):t})}_2}{\norm{\hat{\bsx}_{t+1}}_2}.
\]

\begin{table}[!t]
\centering
\captionof{table}{Hyperparameters in the CViT training.}
\label{tab:hyper-cvit}
\resizebox{\linewidth}{!}{%
\begin{tabular}{lrrrrrr}
\toprule
 & \multicolumn{2}{c}{DR} & \multicolumn{2}{c}{NS} & \multicolumn{2}{c}{SWE}\\
 \cmidrule(lr){2-3} \cmidrule(lr){4-5} \cmidrule(lr){6-7}
Hyperparam. & CViT & X-CViT & CViT & X-CViT & CViT & X-CViT\\
\midrule
Patch Size & (8,8)& (8,8) & (8,8)& (8,8)& (8,8)& (8,8)\\
Grid Size & (128,128)& (128,128)& (128,128)& (128,128)& (96,192)& (96,192)\\
Latent Dim, Embed Dim, \\\quad Depth, Attn Heads & 512,384,5,6& 512,384,6,6& 512,384,5,6& 512,384,5,6& 512,384,5,6& 512,384,5,6\\
Decoder Embed Dim, \\\quad Decoder Attn Heads, \\\quad Decoder Depth & 512,16,1& 512,16,1& 512,16,1& 512,16,1& 512,16,1& 512,16,1\\
Out Dim & 2& 2& 3& 3& 2& 2\\
Input Timesteps & 2& 2& 10& 10& 2& 2\\
Train/Val/Test Splits & 900/0/50& 900/0/50 &6500/0/10& 6500/0/10& 5600/0/10 & 5600/0/10 \\
Minibatch & 16& 32& 16& 16& 32& 32\\
Optim & AdamW & AdamW & AdamW& AdamW& AdamW& AdamW \\
LR Schedule& Warm. Exp. Decay& Warm. Exp. Decay& Warm. Exp. Decay& Warm. Exp. Decay& Warm. Exp. Decay& Warm. Exp. Decay\\
LR init, end, peak & 0, 1E-6, 5E-4& 0, 1E-6, 1E-3& 0, 1E-6, 1E-3 & 0, 1E-6, 1E-3 & 0, 1E-6, 1E-3& 0, 1E-6, 1E-3\\
LR decay, transit, warmup & 0.9, 5000, 5000& 0.9, 5000, 5000& 0.9, 5000, 5000& 0.9, 5000, 5000& 0.9, 5000, 5000& 0.9, 5000, 5000\\
Weight Decay & 1E-5& 1E-5& 1E-5& 1E-5& 1E-5& 1E-5\\
Grad. Clip & 1.0& 1.0& 1.0& 1.0& 1.0& 1.0\\
Iterations & 3E+5& 3E+5& 2E+5& 2E+5& 2E+5& 2E+5 \\
\bottomrule
\end{tabular}%
}
\end{table}
\begin{table}[t]
\centering
\captionof{table}{Hyperparameters in the MPP from-scratch training and pretraining.}
\label{tab:hyper-mpp-fromscratch}
\resizebox{0.7\linewidth}{!}{%
\begin{tabular}{lcccccccccc}
\toprule
& \multicolumn{6}{c}{From-Scratch} & \multicolumn{4}{c}{Pretraining}\\
\cmidrule(lr){2-7} \cmidrule(lr){8-11}
 & \multicolumn{3}{c}{MPP} & \multicolumn{3}{c}{X-MPP} & \multicolumn{2}{c}{MPP} & \multicolumn{2}{c}{X-MPP}\\
 \cmidrule(lr){2-4} \cmidrule(lr){5-7} \cmidrule(lr){8-9} \cmidrule(lr){10-11}
Hyperparam. & DR & SWE & CFD & DR & SWE & CFD & 2D & 1,2D & 2D & 1,2,3D\\
\midrule
Batch Size & 64 & 128 & 8 & 64 & 64 & 8 & 4 & 4 & 16 & 16\\
Input Timesteps & 16 & 16 & 16 & 16 & 16 & 16& 4 & 4 & 4 & 4\\
Accumulation Steps & 5 & 5 & 5 & 5 & 5 & 5 & 5 & 5 & 6 & 5\\
Epochs & 120 & 120 & 120 & 120 & 120 & 120 & 120& 120& 120& 110\\
\bottomrule
\end{tabular}%
}
\end{table}
\begin{table}[!t]
\centering
\captionof{table}{Hyperparameters in the MPP finetuning.}
\label{tab:hyper-mpp-finetuning}
\resizebox{\linewidth}{!}{%
\begin{tabular}{lcccccccccccccccc}
\toprule
& \multicolumn{8}{c}{MPP} & \multicolumn{8}{c}{X-MPP}\\
\cmidrule(lr){2-9} \cmidrule(lr){10-17}
Hyperparam. & DR2D & SWE2D & CFD2D & DS1D & CFD1D & BG1D & ADV1D & CFD3D & DR2D & SWE2D & CFD2D & DS1D & CFD1D & BG1D & ADV1D & CFD3D \\
\midrule
Batch Size & 16& 16& 16& 16& 16& 16& 16& 16& 64& 16& 64& 16& 16& 16& 16& 16\\
Input Timesteps & 4& 4& 4& 4& 4& 4& 4& 4& 4& 4& 4& 4& 4& 4& 4& 4\\
Accumulation Steps & 1& 1& 1& 1& 1& 1& 1& 1& 1& 1& 1& 1& 1& 1& 1& 1\\
Epochs & 120 & 120 & 120 & 120 & 120 & 120 & 120 & 120 & 120 & 120 & 120 & 120 & 120 & 120 & 120 & 120 \\
\bottomrule
\end{tabular}%
}
\end{table}

\textbf{Hyperparameters.}
We report the hyperparameters used for training CViT, X-CViT, MPP, and X-MPP. The hyperparameters include the number of training epochs, train/val/test split, minibatch size, gradient accumulation steps, optimizer, weight decay, drop path probability, learning rate, learning rate scheduling, gradient clipping, and others.

For CViT training, the hyperparameters are listed in~\cref{tab:hyper-cvit}. For MPP training, most hyperparameters are shared across different settings, though some differ. The base values of the hyperparameters are:
\begin{itemize}
    \item \textbf{Epochs:} 120
    \item \textbf{Train/Val/Test Splits:} $X$\%/10\%/10\% split on each dataset at the trajectory level, where $X$ denotes a subsample from 80\% of the total dataset
    \item \textbf{Minibatch Size:} 16
    \item \textbf{Accumulation Steps:} 5
    \item \textbf{Optimizer:} Adan
    \item \textbf{Weight Decay:} 1E-3
    \item \textbf{Drop Path Probability:} 0.1
    \item \textbf{Learning Rate:} DAdaptation
    \item \textbf{Learning Rate Scheduling:} Cosine Decay
    \item \textbf{Gradient Norm Clipping:} 1.0
\end{itemize}
These choices mostly follow the settings of \citet{mccabe2024multiple}. The varying hyperparameters for from-scratch training, pretraining, and finetuning are described in~\cref{tab:hyper-mpp-fromscratch} and~\cref{tab:hyper-mpp-finetuning}.

\section{Partial Differential Equations}
\label{app:sec:pdes}

The PDE solution benchmarks are employed from PDEBench~\citep{takamoto2022pdebench} and PDEArena~\citep{gupta2022towards}. Here are the specifications of the equations and their boundary conditions.

\subsection{2D Shallow Water Equations.} 
The shallow water equations, derived from the general Navier–Stokes equations, provide a reduced-order model for free-surface flows such as waves and dam breaks. In the PDEBench benchmark, they are used to simulate a 2D radial dam break scenario. The governing equations are given by:
\[
\partial_t h + \partial_x (hu) + \partial_y (hv) &= 0, \\
\partial_t (hu) + \partial_x \left(u^2 h + \frac{1}{2} g_r h^2 \right) + \partial_y (uvh) &= -g_r h \partial_x b, \\
\partial_t (hv) + \partial_y \left(v^2 h + \frac{1}{2} g_r h^2 \right) + \partial_x (uvh) &= -g_r h \partial_y b,
\]
where $h$ denotes the water depth, $(u, v)$ are the horizontal velocity components, $g_r$ is the gravitational acceleration, and $b(x, y)$ represents the bathymetry.

The initial condition corresponds to a circular bump in the center of the domain $\Omega = [-2.5, 2.5]^2$:
\[
h(t=0, x, y) = 
\begin{cases}
2.0, & \text{if } \sqrt{x^2 + y^2} < r, \\
1.0, & \text{otherwise},
\end{cases}
\] where the radius $r$ is randomly sampled from the uniform distribution $U(0.3, 0.7)$.

The simulation is performed using the PyClaw finite volume solver. This PDE setup introduces realistic dynamics including shock propagation and wave reflections.

\subsection{2D Compressible Fluid Dynamics.}
The 2D compressible Navier–Stokes equations describe the dynamics of a compressible fluid, accounting for variations in mass, momentum, and energy over time. These equations are fundamental for modeling gas flows where density changes are significant. The system consists of the conservation laws for mass, momentum, and energy:

\[
\label{eq:cfd}
&\partial_t \rho + \nabla \cdot (\rho \mathbf{v}) = 0, \\
&\rho(\partial_t \mathbf{v} + \mathbf{v} \cdot \nabla \mathbf{v}) = -\nabla p + \eta \Delta \mathbf{v} + \left(\zeta + \frac{\eta}{3} \right) \nabla (\nabla \cdot \mathbf{v}), \\
&\partial_t \left( \epsilon + \frac{1}{2} \rho |\mathbf{v}|^2 \right) + \nabla \cdot \left[ \left( \epsilon + p + \frac{1}{2} \rho |\mathbf{v}|^2 \right) \mathbf{v} - \mathbf{v} \cdot \boldsymbol{\sigma}' \right] = 0,
\]

where $\rho$ is the fluid density, $\mathbf{v}$ is the velocity vector, $p$ is the pressure, $\epsilon = p/(\Gamma - 1)$ is the internal energy (with $\Gamma = 5/3$), and $\boldsymbol{\sigma}'$ is the viscous stress tensor. The parameters $\eta$ and $\zeta$ denote the shear and bulk viscosities, respectively.

To generate the data, the simulations employ a second-order accurate HLLC finite volume solver for the inviscid terms, coupled with central differencing for the viscous contributions. This setup enables the benchmark to test model fidelity across a wide range of physically realistic fluid dynamics problems.

\subsection{2D Diffusion-Reaction Equation.}
This equation models the interaction between two spatially distributed quantities (commonly an \textit{activator} and an \textit{inhibitor}) across a two-dimensional domain. It captures complex spatiotemporal behaviors such as wave propagation and pattern formation, often seen in biological or chemical systems.

The system is described by:
\[
\partial_t u &= D_u \left( \partial_{xx} u + \partial_{yy} u \right) + R_u(u, v), \\
\partial_t v &= D_v \left( \partial_{xx} v + \partial_{yy} v \right) + R_v(u, v),
\]
where $ u(t, x, y) $ and $ v(t, x, y) $ denote the concentrations of the activator and inhibitor, respectively. $ D_u $ and $ D_v $ are their diffusion coefficients. The reaction dynamics are governed by the FitzHugh–Nagumo model:
\[
R_u(u, v) = u - u^3 - k - v, \quad R_v(u, v) = u - v,
\]
with $ k = 5 \times 10^{-3} $.

In the benchmark setup, the simulation domain is $ x, y \in (-1, 1) $ and $ t \in (0, 5] $. The initial conditions are generated using Gaussian noise, and Neumann boundary conditions (zero flux) are applied to ensure no flow across domain boundaries. Numerical solutions are computed using the finite volume method with fourth-order Runge–Kutta time integration.

\subsection{2D Incompressible Navier-Stokes Equations (PDEArena).}
The Navier-Stokes equations are a cornerstone of fluid dynamics, describing the motion of fluid substances under the influence of internal and external forces. In PDEArena, the two-dimensional incompressible Navier-Stokes equations are employed to investigate complex multi-scale flow phenomena. These equations govern the evolution of the velocity field \( \mathbf{v}(t, \mathbf{x}) \in \mathbb{R}^2 \) in a domain \( \mathbf{x} \in \mathbb{R}^2 \), and are formulated in the velocity-pressure form as:
\[
\frac{\partial \mathbf{v}}{\partial t} + (\mathbf{v} \cdot \nabla)\mathbf{v} = -\nabla p + \mu \nabla^2 \mathbf{v} + \mathbf{f}, \quad \nabla \cdot \mathbf{v} = 0,
\]
where \( p \) is the pressure, \( \mu \) is the kinematic viscosity (diffusion coefficient), and \( \mathbf{f} \) represents external force, such as buoyancy.

For simulation in PDEArena, an additional scalar field is introduced, representing a passive scalar (e.g., particle concentration) that is advected by the velocity field and interacts with it through an external buoyancy force \( \mathbf{f} = (0, f)^\top \).

The initial conditions include both the velocity and scalar fields, defined over a \( 128 \times 128 \) grid with a resolution of \( \Delta x = \Delta y = 0.25 \). The simulation time-step is \( \Delta t = 1.5 \) seconds, and the domain is closed with Dirichlet boundary conditions \( \mathbf{v} = 0 \) and Neumann conditions \( \partial s / \partial x = 0 \) for the scalar field.

The simulations are numerically solved using the $\Phi$\texttt{Flow} framework and serve as a rich testbed for evaluating neural PDE surrogates in capturing advection-diffusion dynamics, vortex interactions, and response to varying force parameters.

\subsection{Shallow Water Equations (PDEArena).}
The shallow water equations are a set of hyperbolic partial differential equations that describe the flow of a thin layer of incompressible fluid under the influence of gravity. They are derived from the incompressible Navier–Stokes equations by assuming that the horizontal length scales are much larger than the vertical ones, leading to a vertically averaged flow model. In PDEArena, the shallow water equations are used to model both local and global geophysical flow phenomena, such as waves and large-scale atmospheric dynamics.

The equations govern the evolution of the fluid height \( h(t, \mathbf{x}) \) and the horizontal velocity field \( \mathbf{v}(t, \mathbf{x}) = (u, v) \) over a 2D domain \( \mathbf{x} \in \mathbb{R}^2 \), and take the following form:
\[
\begin{aligned}
\frac{\partial h}{\partial t} + \nabla \cdot (h \mathbf{v}) &= 0, \\
\frac{\partial \mathbf{v}}{\partial t} + (\mathbf{v} \cdot \nabla)\mathbf{v} + g \nabla h &= \mu \nabla^2 \mathbf{v} + \mathbf{f},
\end{aligned}
\]
where \( g \) is the gravitational acceleration, \( \mu \) is the viscosity, and \( \mathbf{f} \) represents external forces such as wind stress or Coriolis effects.

The simulations are initialized with spatial fields of velocity and pressure, and are performed on a global grid with resolution \( 192 \times 96 \), corresponding to a spatial discretization of \( \Delta x = 1.875^\circ \), \( \Delta y = 3.75^\circ \), and a temporal resolution of \( \Delta t = 48 \) hours. Periodic boundary conditions are applied in the longitudinal direction, while appropriate boundary conditions (e.g., reflective or free-slip) are used in the latitudinal direction.

These simulations are generated using a modified version of the \texttt{SpeedyWeather.jl} framework. The setup allows for evaluating the performance of neural PDE surrogates on both velocity-pressure and vorticity-stream function formulations, capturing a wide range of scales and flow features relevant to climate and weather modeling.

\subsection{1D Diffusion-Sorption Equation.}
The diffusion-sorption equation models a diffusion process in porous media that is retarded by a non-linear sorption mechanism. This type of process is relevant in real-world applications such as groundwater contaminant transport. In the PDEBench benchmark, it is used to simulate the 1D transport of a solute under the influence of sorption based on the Freundlich isotherm. The governing equation is given by:
\[
\partial_t u(t, x) = \frac{D}{R(u)} \partial_{xx} u(t, x),
\]
where $u(t, x)$ denotes the concentration of the solute, $D = 5 \times 10^{-4}$ is the effective diffusion coefficient, and $R(u)$ is the retardation factor that accounts for the sorption effect. The retardation factor is defined as:
\[
R(u) = 1 + \frac{1 - \phi}{\phi} \rho_s k_{\mathrm{nf}} u^{n_f - 1},
\]
with $\phi = 0.29$ the porosity, $\rho_s = 2880$ the bulk density, $k_{\mathrm{nf}} = 3.5 \times 10^{-4}$ the Freundlich coefficient, and $n_f = 0.874$ the Freundlich exponent.

The initial condition is sampled from a uniform distribution:
\[
u(t=0, x) \sim \mathcal{U}(0, 0.2), \quad \text{for } x \in (0, 1).
\]

The boundary conditions are given by:
\[
u(t, 0) = 1.0, \quad u(t, 1) = D \partial_x u(t, 1),
\]
where the second condition is a Cauchy-type boundary involving a spatial derivative, introducing numerical challenges.

The simulation is performed using a finite volume method for spatial discretization and a fourth-order Runge–Kutta method for time integration. This PDE setup captures realistic nonlinear diffusion behaviors with singularities and complex boundary dynamics.

\subsection{1D Burgers' Equation.}
The Burgers' equation is a fundamental nonlinear partial differential equation that models the interplay between convection and diffusion in fluid dynamics. It serves as a simplified prototype for the Navier–Stokes equations and is used to study shock formation and dissipative processes. In the PDEBench benchmark, the 1D viscous Burgers' equation is used to simulate such nonlinear dynamics. The governing equation is given by:
\[
\partial_t u(t, x) + \partial_x \left( \frac{1}{2} u^2(t, x) \right) = \frac{\nu}{\pi} \partial_{xx} u(t, x),
\]
where $u(t, x)$ is the velocity field and $\nu$ is the diffusion coefficient, representing the kinematic viscosity.

The initial condition is constructed as a superposition of sinusoidal modes:
\[
u(t=0, x) = \sum_{i=1}^{N} A_i \sin(k_i x + \phi_i),
\]
where the wave numbers $k_i = 2\pi n_i / L_x$ are randomly selected integers $n_i \in [1, n_{\text{max}}]$, $A_i \sim \mathcal{U}(0, 1)$ are amplitudes, $\phi_i \sim \mathcal{U}(0, 2\pi)$ are phases, and $L_x = 1$ is the domain length. Additional operations such as applying the absolute value or a window function are applied with small probability to introduce further variability.

The domain is defined as $x \in (0, 1)$, and periodic boundary conditions are imposed:
\[
u(t, 0) = u(t, 1), \quad \partial_x u(t, 0) = \partial_x u(t, 1).
\]

The simulation is performed using a second-order upwind finite difference scheme for the convective term and a central difference scheme for the diffusive term. This PDE setup is particularly suited for studying shock dynamics, nonlinear wave interactions, and the effect of viscosity on solution smoothness.

\subsection{1D Advection Equation.}
The advection equation is a linear hyperbolic partial differential equation that models the transport of a conserved quantity without diffusion or reaction. It serves as a canonical example for studying wave propagation and translation phenomena in physics and engineering. In the PDEBench benchmark, the 1D advection equation is used to simulate pure transport dynamics. The governing equation is given by:
\[
\partial_t u(t, x) + \beta \, \partial_x u(t, x) = 0,
\]
where $u(t, x)$ represents the advected scalar quantity and $\beta$ is the constant advection speed.

The initial condition is defined as a superposition of sinusoidal waves:
\[
u(t=0, x) = \sum_{i=1}^{N} A_i \sin(k_i x + \phi_i),
\]
with $k_i = 2\pi n_i / L_x$ representing the wave numbers for randomly selected integers $n_i \in [1, n_{\text{max}}]$, amplitudes $A_i \sim \mathcal{U}(0, 1)$, phases $\phi_i \sim \mathcal{U}(0, 2\pi)$, and $L_x = 1$ denoting the domain length. With small probability, transformations such as taking the absolute value or applying a window function are applied to the initial field to increase diversity.

The spatial domain is $x \in (0, 1)$, and periodic boundary conditions are employed:
\[
u(t, 0) = u(t, 1), \quad \partial_x u(t, 0) = \partial_x u(t, 1).
\]

The numerical solution is obtained using a second-order upwind finite difference scheme in both space and time. This PDE setup serves as a benchmark for evaluating models' ability to learn and reproduce translational dynamics with minimal distortion or dispersion.

\subsection{1D Compressible Fluid Dynamics.}
The 1D compressible fluid dynamics (CFD) equations model the conservation of mass, momentum, and energy in a compressible fluid. Derived from the general compressible Navier–Stokes equations, they are used to simulate phenomena such as shock waves, rarefaction, and contact discontinuities. In the PDEBench benchmark, this setup includes various configurations such as inviscid flow, viscous flow, and shock-tube initial conditions. The governing equations are given by:
\[
\partial_t \rho + \partial_x (\rho u) = 0, \\
\partial_t (\rho u) + \partial_x \left( \rho u^2 + p \right) = \partial_x \sigma, \\
\partial_t E + \partial_x \left[ (E + p) u \right] = \partial_x (u \sigma),
\]
where $\rho$ is the density, $u$ is the velocity, $p$ is the pressure, $E = \epsilon + \frac{1}{2} \rho u^2$ is the total energy with internal energy $\epsilon = \frac{p}{\Gamma - 1}$, and $\sigma$ is the viscous stress term defined as $\sigma = \eta \partial_x u$ for shear viscosity $\eta$. The ratio of specific heats is set to $\Gamma = 5/3$.

The benchmark includes multiple initial condition types:
\begin{itemize}
\item \textbf{Random field: }Initial $\rho$, $u$, and $p$ fields generated as smooth random perturbations using a superposition of sine waves.
\item \textbf{Shock tube: }A Riemann problem where left and right states $(\rho, u, p)$ are sampled from uniform distributions with a sharp discontinuity at a random position.
\end{itemize}
The spatial domain is $x \in (0, 1)$ with two boundary condition types:
\begin{itemize}
\item \textbf{Periodic: }Fields wrap around the domain,
\item \textbf{Out-going: }Ghost cells copy the nearest interior value to allow waves to exit.
\end{itemize}

The simulations are performed using a second-order HLLC Riemann solver with MUSCL reconstruction for inviscid cases, and central differencing for viscous terms. This PDE setup is challenging due to strong nonlinearity, shock formation, and sensitivity to initial and boundary conditions.

\subsection{3D Compressible Fluid Dynamics.}
The 3D compressible Navier–Stokes equations govern the motion of gases where density, pressure, and velocity fields evolve in space and time. This system models conservation of mass, momentum, and energy in three dimensions, making it essential for simulating realistic high-speed flows, turbulence, and shock dynamics. The equations are same as \cref{eq:cfd}, but their variables are three dimensional, e.g. $ \mathbf{v} \in \mathbb{R}^3 $.

Numerical solutions are generated using a second-order HLLC scheme for the inviscid part and central differencing for viscosity. This 3D setting significantly increases the complexity of the benchmark, introducing challenges in both computational cost and physical realism for surrogate models.


\end{document}